\documentclass[journal]{IEEEtran}
\usepackage{amsthm}
\usepackage{amsfonts}
\usepackage{times}		
\usepackage[cmex10]{amsmath}
\usepackage{amssymb}
\usepackage[switch]{lineno}
\usepackage{lineno}
\usepackage{amsthm}
\usepackage[linesnumbered,ruled]{algorithm2e}
\usepackage{graphicx}
\usepackage{colortbl,booktabs}
\usepackage{threeparttable}
\usepackage{subeqnarray}
\usepackage{cases}
\usepackage{multicol}
\usepackage{multirow}
\usepackage{subfigure}
\usepackage{array}
\usepackage[table]{xcolor}
\usepackage{verbatim}
\usepackage{url}
\newtheorem{theorem}{Theorem}

\newtheorem{definition}{Definition}

\theoremstyle{definition}

\let\oldnl\nl
\newcommand{\nonl}{\renewcommand{\nl}{\let\nl\oldnl}}
\DeclareMathOperator*{\argmin}{argmin}

\ifCLASSINFOpdf
\else
\fi
\begin{document}
%
\title{Effective Image Retrieval via Multilinear Multi-index Fusion}
%
%
%

\author{Zhizhong~Zhang,~
        Yuan~Xie,~\IEEEmembership{Member,~IEEE,}
        Wensheng~Zhang,
        Qi~Tian,~\IEEEmembership{Fellow,~IEEE,}
\thanks{Z. Zhang, Y. Xie, and W. Zhang are with the Research Center of Precision Sensing and Control, Institute of Automation, Chinese Academy of Sciences, Beijing, 100190, China and the School of Computer and Control Engineering, University of Chinese Academy of Sciences, Beijing,
101408, China. E-mail: \{zhangzhizhong2014, yuan.xie\}@ia.ac.cn, zhangwenshengia@hotmail.com}
\thanks{Q. Tian is with the Department of Computer Science, University of Texas at San Antonio, San Antonio, TX 78249 USA; E-mail: qitian@cs.utsa.edu}} 

%
%

\markboth{Journal of \LaTeX\ Class Files,~Vol.~XX, No.~X, June~2017}%
{Shell \MakeLowercase{\textit{et al.}}: Bare Demo of IEEEtran.cls for IEEE Journals}
%



\maketitle


\begin{abstract}
Multi-index fusion has demonstrated impressive performances in retrieval task by integrating different visual representations in a unified framework.
However, previous works mainly consider propagating similarities via neighbor structure, ignoring the high order information among different visual representations. 
In this paper, we propose a new multi-index fusion scheme for image retrieval. By formulating this procedure as a multilinear based optimization problem, the complementary information hidden in different indexes can be explored more thoroughly.
Specially, we first build our multiple indexes from various visual representations. Then a so-called index-specific functional matrix, which aims to propagate similarities, is introduced for updating the original index. The functional matrices are then optimized in a unified tensor space to achieve a refinement, such that the relevant images can be pushed more closer. 
The optimization problem can be efficiently solved by the augmented Lagrangian method with theoretical convergence guarantee. 
Unlike the traditional multi-index fusion scheme, our approach embeds the multi-index subspace structure into the new indexes with sparse constraint, thus it has little additional memory consumption in online query stage. Experimental evaluation on three benchmark datasets reveals that the proposed approach achieves the state-of-the-art performance, {\it i.e.}, N-score 3.94 on UKBench, mAP 94.1\% on Holiday and 62.39\% on Market-1501.
\end{abstract}


%
\IEEEpeerreviewmaketitle

\section{Introduction}
%
%
%
%

\IEEEPARstart{T}{his} paper considers the Content Based Image Retrieval (CBIR), whose aim is to find relevant images in massive visual data. Most CBIR systems are built on various kinds of visual features with different index building methods. It usually consists two steps, where the first step is to describe a image by a vector with fixed dimension, such as the bag-of-visual-words (BOW) \cite{videogoogle}, Fisher vectors \cite{fisher}, Vector of locally aggregated descriptors (VLAD) \cite{VLAD}, and other deep convolutional neural network (CNN) based features \cite{local,deepr};
\begin{figure}[!htb]
  \setlength{\abovecaptionskip}{-5pt}
  \setlength{\belowcaptionskip}{-5pt}
  \centering
    \includegraphics[width=0.48\textwidth]{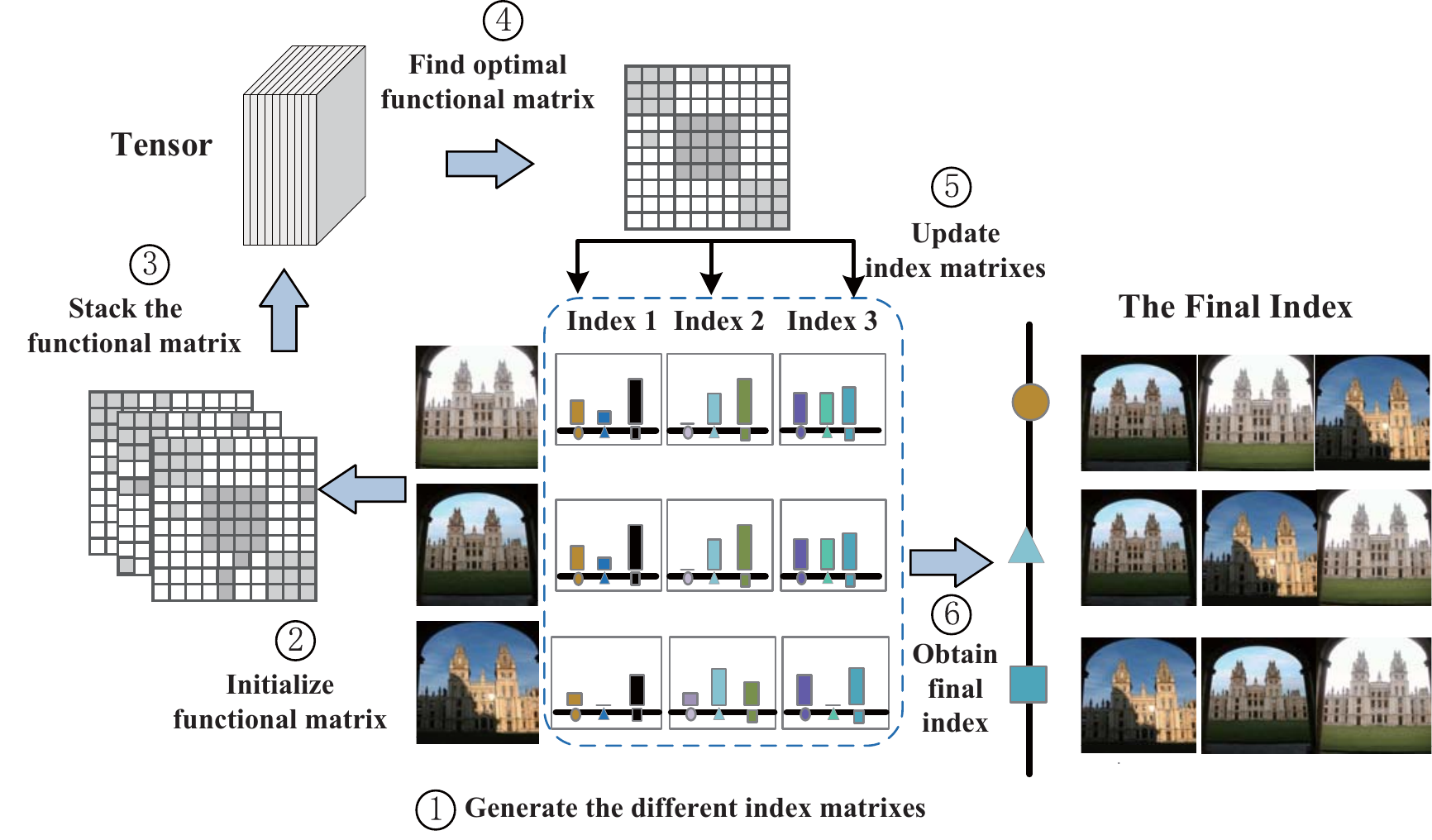}\\
  \caption{The flowchart of the proposed approach.} \label{flowchat} 
\end{figure}
then a simple comparison of two such vectors with cosine distance reflects the similarity of original sets. However, different visual features are different representations of the same instance, which reflects distinct information from different perspectives, {\it e.g.}, SIFT feature has good representative ability for local texture \cite{sift}, while CNN feature focuses on reflecting high level semantic information \cite{alex,vgg}. Although both of these methods are capable of searching visually similar images effectively, totally different results may be obtained, which motivates us to fuse various features \cite{queryfusion,rank,Mulfeature} to boost the retrieval accuracy. But, the feature characteristics and the procedures of index building methods are quite different, such as the holistic feature based method \cite{neuralcodes,binary} and the local feature based method \cite{tembedding,ddaig,uniting}, resulting in the difficulties of fusion on feature level.



Alternatively, a simple yet effective way is to fuse different visual features on index level (also referred to multi-index fusion) \cite{mulllindex,calembed}, which implicitly conduct feature fusion by updating the indexes. 
The index structure is usually considered as a specific database management strategy. By avoiding the exhaustive search, a proper index scheme can significantly promote the efficiency of CBIR system. A representive index structure is the inverted index structure. Local descriptors extracted from the images are firstly quantized to the visual word via nearest neighbor search. Then each image can be indexed as a sparse vector and similar images can be retrieved by counting the co-occurrence of visual words with TF-IDF weighting \cite{videogoogle}. Since only the product of non-zero elements is calculated, inverted index structure has brought the CBIR system to deal with large scale data. Furthermore, the   traditional index building techniques accompanying with deep ConvNet feature have elevated the performance of image retrieval to a new level \cite{bagofconv}.

To make sufficient use of the inverted index structure, previous multi-index fusion works mainly consider propagating similarities via neighbor structure \cite{mulllindex,calembed}. This raises a problem that the high order information among different visual representations is more or less ignored. By contrast, motivated by the multi-view learning methods \cite{LRRR,mulview}, our work learns a index-specific functional matrix to propagate similarities in a unsupervised manner. Instead of simply measuring the Euclidean distance in one visual feature space to find the neighbor structure, our approach optimize the functional matrix in a unified tensor space with the recently proposed tensor-Singular Value Decomposition (t-SVD) based tensor nuclear norm \cite{tsvdnorm}, such that the high order information by comparing every image sample (sample-specific) and every type of visual feature (index-specific) can be captured more effectively and thoroughly.

In this paper, we propose a new multi-index fusion scheme for image retrieval. We formulate this procedure as a multilinear based optimization problem to find a index-specific functional matrix. 
We need to emphasize here that our contributions are not meant as a simple combination between \cite{calembed} and \cite{mulview}. The proposed method (called as MMF) carefully considers the sparse index structure for retrieval, which is the intrinsic property of inverted index structure. Meanwhile, the complementary information captured by high order tensor norm can be propagated via the index-specific functional matrix. Although the proposed method seems to need an unaffordable computing cost and memory usage, the heavy procedure is performed offline only once at training time and can be further invested by dividing images into groups. In summary, the key insight of our approach is to propagate similarity via high order (tensor) information in a unsupervised manner, which implicitly conduct feature fusion on index level. Fig. \ref{flowchat} shows the pipeline of our proposed scheme.

The main contributions of this paper are summarized as follows:
\begin{itemize}
\item We propose a new multi-index fusion scheme to implicitly conduct feature fusion on index level, where complementary information from all visual indexes can be effectively explored via high-order low-rank tensor norm.
\item We present an efficient optimization algorithm to solve the proposed objective function, with relatively low computational complexity and theoretical convergence guarantee.
\item We conduct the extensive evaluation of our method on several challenge datasets, where a significant improvement over the state-of-the-art approaches is achieved. By regarding  person re-identification as a special retrieval task, the proposed model has achieved highly competent (even better) performance compared to recent proposed method.
\end{itemize}

The rest of this paper is organized as follows. Section \ref{relatework} introduces related works. Section \ref{Preliminaries} gives the notations that will be used throughout the paper and the preliminaries on tensors. In Section \ref{proposed}, we review previous multi-index fusion method and motivate our model in detail, give an optimization algorithm to solve it, and analyze its convergence. In Section \ref{Experiment}, we show our experimental analysis and completion results to verify our method. Then we analyse and discuss the proposed model in detail. Finally, we conclude the proposed method in Section \ref{conclusion}.



\section{Related work}\label{relatework}
Most of the CBIR systems can be roughly divided into two parts: image representation and image indexing. Additionally, our work is also related with the multi-feature fusion and multi-view subspace learning. Their strengths and limitations are briefly reviewed below.
\subsection{Image representation}
Image representation has been extensively studied in recent years. To give more discriminative description for image, local features such as SIFT \cite{sift} are introduced in CBIR systems \cite{videogoogle}. Due to its good property of invariance to orientation, uniform scaling and illumination changes, BOW based CBIR systems achieve great success \cite{haming,ukbench,three}. During this period, several methods are proposed to promote the discrimination of BOW based image representation, such as the Hamming embedding \cite{haming}, negative evidence \cite{negative}, soft assignment \cite{soft} and so on.

Meanwhile, a lot of works aim to produce the compact image representation \cite{tembedding,ddaig,VLAD,fisher}, which is benefit for computational efficiency and memory cost. Furthermore, several recent proposed methods attempt to extract features from the pre-trained deep convolutional networks via compact encoding. By using the compact codes, Babenko {\it et al.} discover that the features from the fully-connected layers of CNN (fully-connected feature) provide high-level descriptors of the visual content \cite{neuralcodes}, yielding competitive results. But more recently, the research attention has moved to the activations of CNN filters (convolutional feature) \cite{local}. Convolutional features have a natural interpretation as descriptors of local image regions, which not only share the same benefits with the local features, but also hold high-level semantic information \cite{convfeature}. Empirically, they gain even better results than the local features. Generally speaking, both of these methods hold distinct merits, resulting in different retrieval results. This may cause us to consider whether we should only focus on one type visual feature ({\it e.g.,} abandon these hand-crafted features), or combine different visual representations for retrieval.

\subsection{Image indexing}
Indexing local features by inverted index structure and hashing holistic features by compact binary codes have been two mainstreams methods in recent years. For the hash technique, data-independent hash method can produce high collision probability, but often needs long hash bits and multiple hash tables \cite{innerproduct,LSH}. Data-dependent hash methods, such as Stochastic Multiview Hashing \cite{mulhash}, Spectral Hashing \cite{spectralhash} and Nonlinear Sparse Hashing \cite{NSH} 
aim to generate short binary codes via a learning processing, which is more effectively and efficiently. We refer the readers to \cite{hashreview} for a comprehensive review. Although it provides accurate search results, the hash method is a method that loss information. In contrast, inverted index structure, as one of lossless indexing methods, is prevalently utilized in the BOW based image search, which has shown excellent scalability by extensively studies \cite{haming}.

For inverted index structure, previous works mainly focus on adding detail information into the inverted indexes after the seminal work \cite{videogoogle}. Zhou {\it et al}. \cite{spatial} index the geometric clues of local features via spatial coding. Zhang {\it et al.} \cite{coindexing} jointly embed the local features and semantic clues into the inverted indexes. Babenko {\it et al.} propose a inverted multi-index framework to reduce the quantization loss \cite{mulindex}. Recently, Mohedano {\it et al.} encode the convolutional features via Bag-of-words scheme, where competitive results demonstrate the suitability of the BOW based index building methods for CNN features \cite{bagofconv}. 
\subsection{Feature fusion}
To take full advantage of the strengths of each feature, a lot of works have already begun to combine different visual features to boost the retrieval performance \cite{queryfusion,rank,coupled}. In \cite{rank}, Zhang {\it et al.} conduct the fusion in ranking stage. By performing a link analysis on a fused graph, the retrieval accuracy can be greatly improved. Zheng {\it et al.} \cite{queryfusion} introduce a score-level fusion method for similar image search. Zheng {\it et al.} \cite{coupled} propose a coupled Multi-index framework to conduct the feature fusion. Nevertheless, these methods treat each image representation independently, ignoring the complementarity among different visual features. Moreover, query operations must be performed multiple times for multiple indexes.

To overcome these drawbacks, some works focus on fusing visual features on index level. 
A common assumption shared in these methods is that: two images, which are nearest neighbors to each other under one type of visual representation, are probably to be true related. By pushing them closer in other visual feature spaces, the search accuracy can be greatly promoted. Under the guidance of this principle, the proposed collaborative index embedding method \cite{calembed}, which is most relevant to our work, utilize an alternating index update scheme to fuse feature. By enriching the corresponding feature, it refine the neighborhood structures to improve the retrieval accuracy. Chen {\it et al.} \cite{mulllindex} extend this model for the multi-index fusion problem. 
However, both of these methods neglect the distance information of original feature space. More importantly, high order information is more or less ignored.
\subsection{multi-view subspace learning}
Our work is also related with the multi-view subspace learning methods, especially the subspace clustering methods. 
Sparse subspace clustering \cite{SSC} and low-rank representation \cite{LRRR} are most popular subspace clustering methods, which explore the relationships between samples via self-representation matrix. 
Zhang {\it et al.} \cite{msc} extend the low-rank representation to the multi-view setting via imposing a unfolding high-order norm to the subspace coefficient tensor. However, this tensor constraint can not explore the complementary information thoroughly, due to the fact that the low-rank norm penalize each view equally. 
By using a new tensor construction method, Xie {\it et al.} \cite{mulview} replace the unfolding tensor norm with a recently proposed t-SVD based norm \cite{tsvdnorm,ts}, which is based on a new tensor computational framework \cite{tproduct}. This framework provides a closed multiplication operation between tensors \cite{tsvd}, where the familiar tools used in matrix case can be directly extended to tensor case. Hence, it has good theoretical properties for handling complicated relationship among different views. 
For more detail information, we refer readers to read the section \ref{Preliminaries}.
\section{Background and Preliminaries}\label{Preliminaries}
In this section, we will introduce the notations and basic concepts used in this paper.
\subsection{Basic Notations}
\begin{figure}[!htb]
  \setlength{\abovecaptionskip}{-5pt}
  \setlength{\belowcaptionskip}{-5pt}
  \centering
    \includegraphics[width=0.45\textwidth]{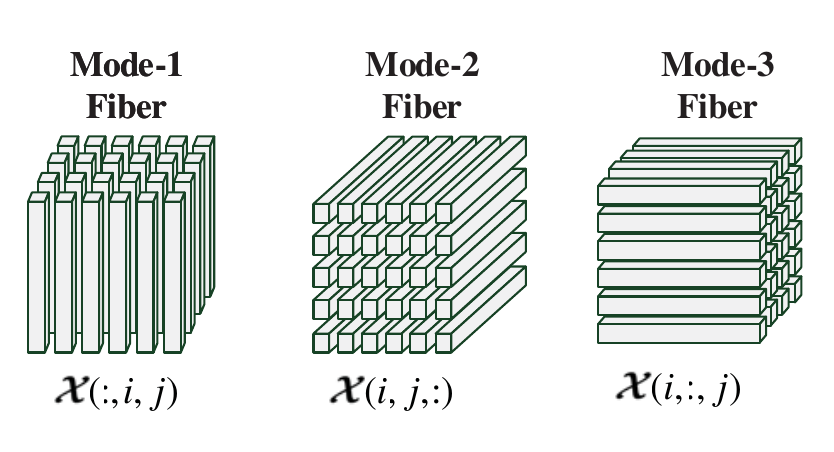}\\
  \vspace{-0.25in}
    \includegraphics[width=0.45\textwidth]{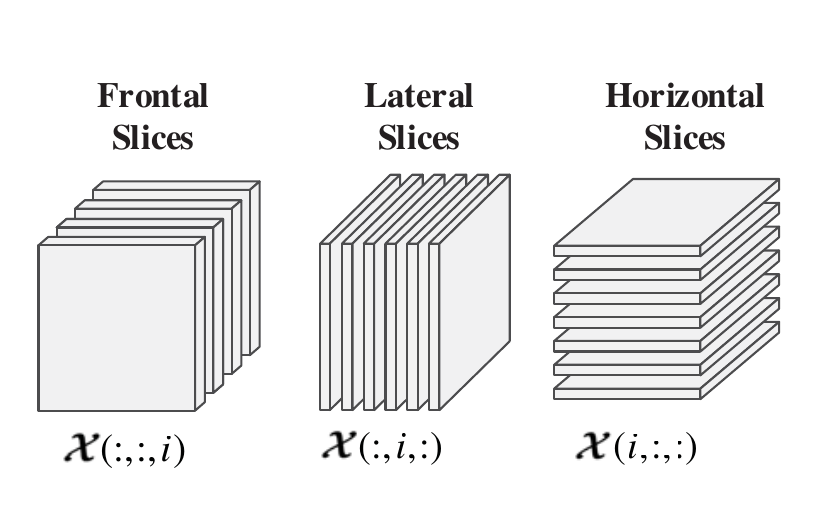}
  \caption{The 1D section and 2D section of a 3-order tensor.} \label{fiberslice} 
\end{figure}

We use bold lower case letters $\mathbf{x}$ to denote vector ({\it e.g.,} BOW based sparse histogram), bold upper case letters $\mathbf{X}$ to denote matrix, and lower case letters $x_{ij}$ for entries of matrix. The notation $\|\mathbf{X}\|_F:=({\sum_{i,j}{|x_{ij}|^2}})^{\frac{1}{2}}$ , $\|\mathbf{X}\|_{2,1}:=\sum_i({\sum_j{x_{ij}^2}})^{\frac{1}{2}}$ and $\|\mathbf{X}\|_{1}:=\sum_{i,j}{|x_{ij}|}$ are the Frobenius norm, the ${l}_{2,1}$-norm and the ${l}_{1}$-norm for matrix, respectively. $\|\mathbf{X}\|_*:=\sum_i\sigma_i(\mathbf{X})$ is the matrix nuclear norm, where $\sigma_i(\mathbf{X})$ denotes the $i$-th largest singular value of a matrix.
The bold calligraphy letters are denoted for tensors ({\it i.e.,} $\boldsymbol{\mathcal{Z}} \in \mathcal{R}^{n_1\times n_2\times n_3}$ is a three-order tensor, where order means the number of ways of the tensor and is fixed at 3 in this paper). For a three-order tensor $\boldsymbol{\mathcal{X}}$, the 2D section $\boldsymbol{\mathcal{X}}(i,:,:)$, $\boldsymbol{\mathcal{X}}(:,i,:)$ and $\boldsymbol{\mathcal{X}}(:,:,i)$ (Matlab notation is used for better understanding) denote the $i$th horizontal, lateral and frontal slices. Analogously, the 1D section $\boldsymbol{\mathcal{X}}(i, j, :)$, $\boldsymbol{\mathcal{X}}(i, :, j)$ and $\boldsymbol{\mathcal{X}}(:, i, j)$ are the mode-1, mode-2 and mode-3 fibers of tensor, as shown in Fig. \ref{fiberslice}. Specially, $\boldsymbol{\mathcal{X}}^{(k)}$ is used to represent $k$th Frontal Slice ${\boldsymbol{\mathcal{X}}}(:, :, k)$ for convenience. And $\boldsymbol{\mathcal{X}}_{f}$ denotes the tensor that we apply Fourier transform to $\boldsymbol{\mathcal{X}}$ along the third dimension.

\subsection{t-SVD framework and key results}
Before we introduce the t-SVD based tensor nuclear norm (TNN-norm), there is a need to give some pre-definitions about the new computational framework \cite{tsvd,tproduct} for a better interpretation.
\begin{definition}[\textbf{t-product}]\label{def:t-prod}
Let $\boldsymbol{{\mathcal{X}}} \in \mathcal{R}^{n_{1} \times n_{2} \times n_{3}}$ and $\boldsymbol{{\mathcal{Y}}} \in \mathcal{R}^{n_{2} \times n_{4} \times n_{3}}$ be tensors. Then the t-product $\boldsymbol{{\mathcal{M}}}=\boldsymbol{{\mathcal{X}}}*\boldsymbol{{\mathcal{Y}}}$ is an $n_{1} \times n_{4} \times n_{3}$ tensor defined as£º
\begin{equation}\label{tpro}
\left[
\begin{matrix}
 \boldsymbol{{\mathcal{M}}}^{(1)} \\
 \boldsymbol{{\mathcal{M}}}^{(2)} \\
 \vdots         \\
 \boldsymbol{{\mathcal{M}}}^{(n_{3})}
\end{matrix}
\right] =
\left[
\begin{matrix}
 \boldsymbol{{\mathcal{X}}}^{(1)}    &\boldsymbol{{\mathcal{X}}}^{(n_{3})}  & \cdots         & \boldsymbol{{\mathcal{X}}}^{(2)} \\
 \boldsymbol{{\mathcal{X}}}^{(2)}    &\boldsymbol{{\mathcal{X}}}^{(1)}      &  \cdots        & \boldsymbol{{\mathcal{X}}}^{(3)} \\
 \vdots            &\ddots              & \ddots         & \vdots         \\
 \boldsymbol{{\mathcal{X}}}^{(n_{3})}&\boldsymbol{{\mathcal{X}}}^{(n_{3}-1)}& \cdots         & \boldsymbol{{\mathcal{X}}}^{(1)}
\end{matrix}
\right]\cdot\left[
\begin{matrix}
 \boldsymbol{{\mathcal{Y}}}^{(1)} \\
 \boldsymbol{{\mathcal{Y}}}^{(2)} \\
 \vdots         \\
 \boldsymbol{{\mathcal{Y}}}^{(n_{3})}
\end{matrix}
\right]
\end{equation}
where $\cdot$ is the standard matrix multiplication.
\end{definition}

\begin{definition}[\textbf{Transpose}]
\label{def:t-trans}
If $\boldsymbol{{\mathcal{X}}} \in \mathcal{R}^{ n_{1} \times n_{2} \times n_{3} }$, then the $\boldsymbol{{\mathcal{X}}}^{\mathrm{T}}$ is an $n_{2} \times n_{1} \times n_{3}$ tensor by transposing each frontal slice of $\boldsymbol{{\mathcal{X}}}$ and reversing the order of the transposed frontal slices 2 through $n_3$.
\end{definition}
\begin{definition}[\textbf{Orthogonal}]\label{rere}
A tensor $\boldsymbol{{\mathcal{Q}}} \in \mathcal{R}^{n_{1} \times n_{1} \times n_{3}}$ is orthogonal if
\begin{equation}
\boldsymbol{{\mathcal{Q}}}^{\mathrm{T}}*\boldsymbol{{\mathcal{Q}}} = \boldsymbol{{\mathcal{Q}}}*\boldsymbol{{\mathcal{Q}}}^{\mathrm{T}} = \boldsymbol{{\mathcal{I}}},
\end{equation}
where $\boldsymbol{{\mathcal{I}}}\in \mathcal{R}^{ n_{1} \times n_{1} \times n_{3} }$ is the identity tensor whose first frontal slice is the identity matrix and other frontal slices are zero.
\end{definition}

Based on the above definitions, it is easy to obtain that t-product can be transformed to matrix multiplication of frontal slices in the Fourier domain. Formally Eq. (\ref{tpro}) equals to:
\begin{equation}
\boldsymbol{{\mathcal{M}}}_{f}^{(k)} = \boldsymbol{{\mathcal{X}}}_{f}^{(k)} \boldsymbol{{\mathcal{Y}}}_{f}^{(k)},~ k = 1,\ldots,n_{3},
\end{equation}
Thus t-product can be calculated efficiently via Fourier transform. And more importantly, an important theoretical resulting property \cite{tsvd} can be concluded from the t-product framework, which is similar to matrix case.
\begin{theorem}[\textbf{t-SVD}]\label{tsvd}
Let $\boldsymbol{{\mathcal{X}}} \in \mathcal{R}^{n_{1} \times n_{2} \times n_{3}}$ be a real-valued tensor. Then  $\boldsymbol{{\mathcal{X}}}$ can be decomposed as£º
\begin{equation}\label{fml:t-svd}
\boldsymbol{{\mathcal{X}}}=\boldsymbol{{\mathcal{U}}}*\boldsymbol{{\mathcal{S}}}*\boldsymbol{{\mathcal{V}}}^{\mathrm{T}},
\end{equation}
where $\boldsymbol{{\mathcal{U}}}\in \mathcal{R}^{n_{1} \times n_{1} \times n_{3}} $ and $\boldsymbol{{\mathcal{V}}}\in \mathcal{R}^{n_{2} \times n_{2} \times n_{3}}$ are orthogonal
tensors. $\boldsymbol{{\mathcal{S}}}$ is an  ${n_{1} \times n_{2} \times n_{3}}$ tensor whose each frontal slices is diagonal matrix.
\end{theorem}
Theorem \ref{tsvd} tells us that any real-valued tensor can be written as the t-product of tensors, which is analogous to matrix SVD. Meanwhile, its derived equivalence Eq. (\ref{fml:t-svd}) in the Fourier
domain can be given as :
\begin{equation}
\begin{aligned}
\label{fml:bdiag}
\left[
\begin{matrix}
{\boldsymbol{\mathcal{X}}}^{(1)}_f & & \\
 &\ddots & \\
 && {\boldsymbol{\mathcal{X}}}^{(n_{3})}_f
\end{matrix}
\right] =
\left[
\begin{matrix}
{\boldsymbol{\mathcal{U}}}^{(1)}_f & & \\
 &\ddots & \\
 && {\boldsymbol{\mathcal{U}}}^{(n_{3})}_f
\end{matrix}
\right] \cdot \\
\left[
\begin{matrix}
{\boldsymbol{\mathcal{S}}}^{(1)}_f & & \\
 &\ddots & \\
 && {\boldsymbol{\mathcal{S}}}^{(n_{3})}_f
\end{matrix}
\right]\cdot
\left[
\begin{matrix}
{\boldsymbol{\mathcal{V}}}^{(1)}_f & & \\
 &\ddots & \\
 && {\boldsymbol{\mathcal{V}}}^{(n_{3})}_f
\end{matrix}
\right]^{\mathrm{T}},
\end{aligned}
\end{equation}
where $\boldsymbol{\mathcal{X}}^{(i)}_f = \boldsymbol{\mathcal{U}}^{(i)}_f \boldsymbol{\mathcal{S}}^{(i)}_f (\boldsymbol{\mathcal{V}}^{(i)}_f)^{\mathrm{T}}, i=1,\ldots,n_3$ are standard matrix SVD.
Thus the t-SVD based tensor nuclear norm \cite{ts} is given as
\begin{equation}
\label{fml:gtnn}
||\boldsymbol{\mathcal{X}}||_{\circledast} :=\sum_{i=1}^{\mathrm{min}(n_{1},n_{2})}\sum_{k=1}^{n_{3}}|{ \boldsymbol{\mathcal{S}}}_{f}(i,i,k)|.
\end{equation}
Due to the fact that the diagonal block matrix in Fourier domain can be reversed to cyclic matrix in origin domain \cite{tsvd}, t-SVD based tensor nuclear norm can be also given as:
\begin{equation}\label{circ}
||\boldsymbol{\mathcal{X}}||_{\circledast}=\|\left[
\begin{matrix}
 \boldsymbol{{\mathcal{X}}}^{(1)}    &\boldsymbol{{\mathcal{X}}}^{(n_{3})}  & \cdots         & \boldsymbol{{\mathcal{X}}}^{(2)} \\
 \boldsymbol{{\mathcal{X}}}^{(2)}    &\boldsymbol{{\mathcal{X}}}^{(1)}      &  \cdots        & \boldsymbol{{\mathcal{X}}}^{(n_3)} \\
 \vdots            &\ddots              & \ddots         & \vdots         \\
 \boldsymbol{{\mathcal{X}}}^{(n_{3})}&\boldsymbol{{\mathcal{X}}}^{(n_{3}-1)}& \cdots         & \boldsymbol{{\mathcal{X}}}^{(1)}
\end{matrix}
\right]\|_{*}
\end{equation}
Different from the tensor nuclear norm (sum the matrix nuclear norm of unfolding matrix of tensor) defined in \cite{msc}, TNN-norm measures the tensor rank by comparing every row and every column of each frontal slices, which is the tightest convex approximation to $l_1$ norm of tensor multi-rank proved by \cite{ts}.


\section{The proposed Methods}\label{proposed}

Multi-index fusion is a technique for implicitly conducting feature fusion on index level, where we can only keep one visual index for both effect and efficient image retrieval. Suppose we have $V$ types of feature indexes denoted as $\mathbf{X}_1,\mathbf{X}_2,\ldots,\mathbf{X}_V\in{\mathcal{R}^{d_v\times N}}$, whose column is a feature vector ({\it e.g.,} BOW based histograms), $d_v$ is the dimension of $v$-th visual representation, and $N$ is the image number in the database.

Previous multi-index fusion strategies \cite{mulllindex,calembed} mainly consider propagating similarity via the neighbor structure through different indexes.
As suggested in \cite{calembed}, feature fusion on index level can be formulated as:
\begin{align}\label{index-update}
\mathbf{\widetilde{X}}_1=\mathbf{X}_1+\alpha\cdot g(\mathbf{X}_1)\odot \mathbf{X}_1\mathbf{\Phi}_2,\\ \notag
\mathbf{\widetilde{X}}_2=\mathbf{X}_2+\beta\cdot g(\mathbf{X}_2)\odot \mathbf{X}_2\mathbf{\Phi}_1,
\end{align}
where $\alpha$ and $\beta$ are constant factors, $g(\cdot)$ is a zero-indicator function equaling 1 if the element is zero. And $\odot$ denotes the element-wise multiplication operator. The $\mathbf{\Phi}_m,~m=1,2$ is defined as:
\begin{equation}\label{phi}
    \left\{
        \begin{aligned}
     &\mathbf{\Phi}_m(k,i)=1,~\text{if}~ k\neq i,\mathbf{x}_k\in \mathcal{R}_m(\mathbf{x}_i)\\
     &\mathbf{\Phi}_m(k,i)=0,~\text{otherwise}
        \end{aligned}
    \right.\\
\end{equation}
where $\mathcal{R}_m(\mathbf{x}_i),~m=1,2$ denote the neighbor sets of image $i$ in feature index $m$. Eq. (\ref{index-update}) and Eq. (\ref{phi}) assume that the neighbor information in one feature space need to be embedded into the updated feature vector in another feature space, such that the distance between similar images will be reduced. However this method can only handle two feature fusion problem. 

To keep their own characteristics of each visual index, we learn a index-specific functional matrix for updating the index matrices instead. 
We formulate the index updating scheme as:
\begin{align}\label{updatematrix}
\mathbf{X}^*_v=\mathbf{X}_v(\mathbf{I}+\mathbf{Z}_v),~v=1,2,\ldots,V
\end{align}
 \begin{figure*}[!htb]
  \setlength{\abovecaptionskip}{-5pt}
  \setlength{\belowcaptionskip}{-5pt}
  \centering
    \includegraphics[width=0.95\textwidth]{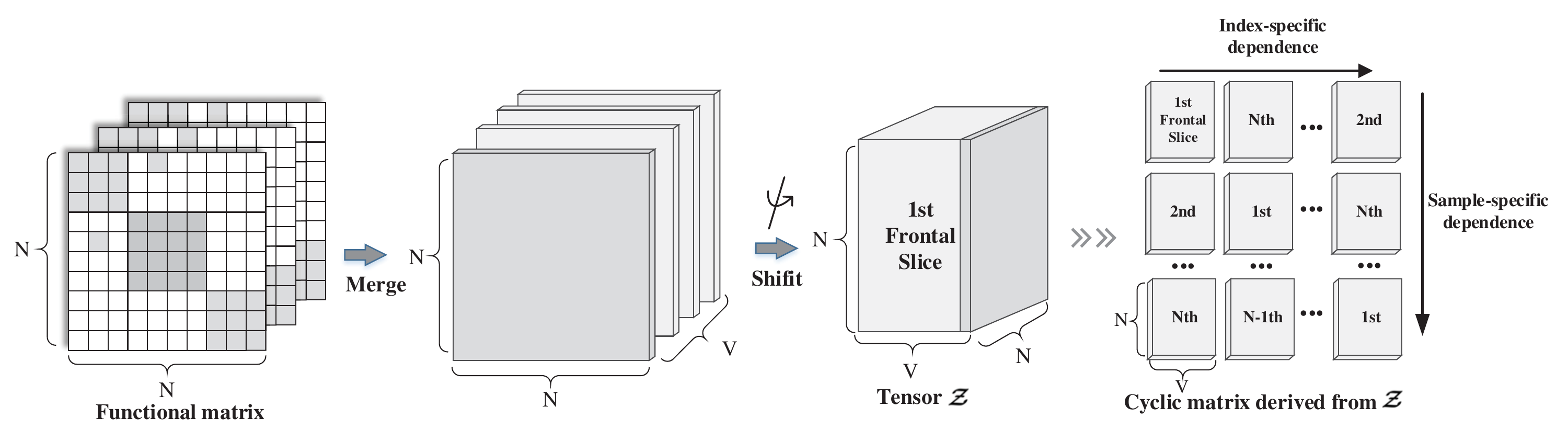}\\
  \caption{The construction of tensor $\boldsymbol{\mathcal{Z}}$ and its derived cyclic matrix.} \label{shift-tensor} 
\end{figure*}
where $\mathbf{I}$ is the identity matrix, $\mathbf{Z}_v\in\mathcal{R}^{N\times N}$ is named as functional matrix and optimized in the unified tensor space in this paper, whose element $z^{v}(i,j)$ is greater than $0$, meaning that image $i$ and image $j$ should be pushed closer. Flowchart 3 to 5 of Fig. \ref{flowchat} show the learning procedure. This fusion procedure can be regarded as an "off-line" query-expansion. Given the functional matrices, the similarities in different visual indexes can be propagated through different visual representations, such that the index matrices can achieve a refinement with much more discriminative power. 
In the following, we will introduce the new scheme to obtain the functional matrix $\mathbf{Z}_v$.

\subsection{Motivation}
There are two basic assumptions in this work, which clearly illustrates our motivation. One is assuming that the related images can be interpreted as a special subspace structure by regarding the gallery as the whole space. This assumption is based on the observation that the corresponding feature vectors of related images are similar to each other, which is analogous to subspace structure. 
We call this assumption as sample-specific dependence. The another is the index-specific dependence, which assumes that the similarities between images measured in different feature space are highly dependent. As discussed above, different visual representations hold distinct merits and thus their search results differ a lot, but that does not mean they have nothing in common. To put it simply, related images are still close among most feature spaces, even if they may not be the nearest neighbors in minority indexes. That is to say, what we need to learn is the consistency rather than the diversity. 

\subsection{multi-linear based multi-index fusion}
In this paper, we utilize a multilinear based optimization to model both dependence. Specially, we consider the self-representation based method \cite{LRRR,mulview}. Formally, we present our model as follow:
\begin{align}\label{ourmodel}
     &\min_{\mathbf{Z}_v,\mathbf{E}}\lambda\|\mathbf{E}\|_{2,1}+ \|\boldsymbol{ \mathcal{Z}}\|_{\circledast}+\sigma\sum_v\|\mathbf{Z}_v\|_1 \\ \notag
     &\text{s.t.} \quad \mathbf{X}_{v}=\mathbf{X}_{v}\mathbf{Z}_{v}+\mathbf{E}_{v}, v=1,2,\ldots,V,
\end{align}
where $\sigma$ and $\lambda$ are constant parameters to control the reconstruction errors and the sparseness of functional matrix, respectively. $\mathbf{X}_v \in \mathcal{R}^{d\times{N}}$ denotes the $v$-th index matrix, and $d$ is the dimension of feature vector, $\mathbf{Z}_v \in \mathcal{R}^{N\times{N}}$ denotes the $v$-th index-specific functional matrix. $\boldsymbol{\mathcal{Z}}=\Phi(\mathbf{Z}_1,\mathbf{Z}_2,\ldots,\mathbf{Z}_V)\in \mathcal{R}^{N\times V \times N}$ is a tensor by merging different $\boldsymbol{\mathbf{Z}}_{v}$ to a 3-order tensor and then shifting illustrated in Fig. \ref{shift-tensor}, $\mathbf{E}=[\mathbf{E}_1,\mathbf{E}_2,\ldots,\mathbf{E}_V]$ is the error matrix, $\|\cdot\|_{\circledast}$ denotes the t-SVD based tensor nuclear norm.

Consequently, $\|\mathbf{E}\|_{2,1}$ in Eq. (\ref{ourmodel}) attempts to control the reconstruction errors, whose aim is to update the index matrices mildly and keep the original representation of database images. The t-SVD based norm $\|\boldsymbol{\mathcal{Z}}\|_{\circledast}$ is used for exploring both dependence by comparing every row (sample-specific) and every column (index-specific) from Eq. (\ref{circ}). For the sample-specific dependence, we assume each functional matrix $\boldsymbol{\mathbf{Z}}_v$ has low rank property. While for the index-specific dependence, we use the high correlations of the index-specific functional matrices for a replacement, where we also assume the different functional matrices share the low rank structure. As a result, related images' information can be embedded into the "new" feature vector via the updating scheme. 
Meanwhile, the sparse constraint $\|\mathbf{Z}_v\|_1$ aims to embed the most significant relevant images into the new index and keep the sparseness of the updated indexes.

\subsection{Optimization Procedure} \label{Optimization}
We can use the Augmented Lagrange Multiplier (ALM) \cite{alm} to solve this optimization problem efficiently. By introducing the auxiliary tensor $\boldsymbol{\mathcal{G}}$ and the auxiliary matrices $\mathbf{M}_v,~v=1,2,\ldots,V$, the optimization problem can be transferred to:
\begin{align}\label{unconstrained}
    &\mathcal{L}(\mathbf{Z}_{1},\ldots,\mathbf{Z}_{V};\mathbf{E}_{1},\ldots,\mathbf{E}_{V};\mathbf{M}_{1},\ldots,\mathbf{M}_{V};\boldsymbol{\mathcal{G}})\notag \\
    &=\sum_v (\sigma \|{\mathbf{M}_v}\|_1+\left\langle {{{\bf{Y}}_v},{{\bf{X}}_v} - {{\bf{X}}_v}{{\bf{Z}}_v} - {{\bf{E}}_v}} \right\rangle \notag \\
    &+\frac{\mu }{2}||{{\bf{X}}_v} - {{\bf{X}}_v}{{\bf{Z}}_v}- {{\bf{E}}_v}||_F^2{\rm{ + }} \left\langle {{{\bf{N}}_v},{{\bf{Z}}_v} - {{\bf{M}}_v}} \right\rangle \notag \\
     &+ \frac{\xi }{2}||\mathbf{Z}_v - \mathbf{M}_v||_F^2 ) +\lambda\|\mathbf{E}\|_{2,1}+\|\boldsymbol{\mathcal{G}}\|_{\circledast} \notag \\
    &+\left\langle {{{\boldsymbol{\cal W}}},{{\boldsymbol{\cal Z}}} - {\boldsymbol{\cal G}}} \right\rangle  + \frac{\rho }{2}||{{\boldsymbol{\cal Z}}} - {\boldsymbol{\cal G}}||_F^2,
\end{align}
where the matrix $\mathbf{N}_v$,$\mathbf{Y}_v$, and the tensor $\boldsymbol{\cal W}$ are Lagrange multipliers. $\mu$, $\xi$ and $\rho$ are the penalty parameters. An accurate and joint optimization of  $\mathbf{E}_v$, $\mathbf{M}_v$, $\mathbf{Z}_v$ and $\boldsymbol{\mathcal{G}}$ seems to be costly. In contrast, we adopt an alternating scheme and partition the unconstrained problem into four steps alternatingly. \\
\textbf{ Subproblem $\mathbf{Z}_v$}: When the $\boldsymbol{\mathcal{G}}$, $\mathbf{E}$, $\mathbf{M}$ are fixed, we will solve the following subproblem for updating the functional matrix $\mathbf{Z}_v$:
\begin{align}
&\mathop {\min }\limits_{{{\bf{Z}}_v}} \left\langle {{{\bf{Y}}_v},{{\bf{X}}_v} - {{\bf{X}}_v}{{\bf{Z}}_v} - {{\bf{E}}_v}} \right\rangle+ \frac{\mu }{2}||{{\bf{X}}_v}- {{\bf{X}}_v}{{\bf{Z}}_v}  \notag \\
& - {{\bf{E}}_v}||_F^2 +\langle {{{\bf{N}}_v},{{\bf{Z}}_v} - {{\bf{M}}_v}}\rangle+ \frac{\xi}{2}||{{\bf{Z}}_v}- {{\bf{M}}_v}||_F^2 \rangle \notag  \\
& + \langle {{{\bf{W}}_v},{{\bf{Z}}_v}{\rm{ }} - {\rm{ }}{{\bf{G}}_v}} \rangle  +\frac{\rho }{2}||{{\bf{Z}}_v}{\rm{ }} - {\rm{ }}{{\bf{G}}_v}||_F^2.
\end{align}
It is easy to solve this optimization problem due to the closed-form solution. We can obtain the solution by setting the derivative to 0:
\begin{align}\label{ZV}
{{\bf{Z}}^*_v} = &( {{\bf{X}}_v}^T{{\bf{Y}}_v} + \mu {{\bf{X}}_v}^T{{\bf{X}}_v} - \mu {{\bf{X}}_v}^T{{\bf{E}}_v} - {{\bf{W}}_v} - {{\bf{N}}_v} \notag \\\nonumber
\\&+\rho {{\bf{G}}_v} + \xi {{\bf{M}}_v})/(\rho  + \xi ) {( {{\bf{I}} + \frac{\mu }{{\rho  + \xi }}{{\bf{X}}_v}^T{{\bf{X}}_v}} )^{ - 1}}.
\end{align}\\
\textbf{ Subproblem $\mathbf{M}_v$}: When $\boldsymbol{\mathcal{G}}$, $\mathbf{E}$, $\mathbf{Z}$ are fixed, solving Eq. (\ref{unconstrained}) is equal to minimize the following problem for updating the auxiliary matrix $\mathbf{M}_v$:
\begin{align}\label{SoftTh}
\min_{\mathbf{M}_v}\sigma \left\| {{{\bf{M}}_v}} \right\|_1 + \frac{\xi }{2}||{{\bf{M}}_v} - ( {{{\bf{Z}}_v} + \frac{1}{\xi }{{\bf{N}}_v}} )||_F^2.
\end{align}\
We can use Soft Thresholding to solve Eq. (\ref{SoftTh}).
\begin{align}\label{mv}
\mathbf{M}^*_v(i,j)=&sign(\mathbf{Z}_v(i,j)+\frac{1}{\xi}{\mathbf{N}}_v(i,j))\cdot \notag\\
&max(|\mathbf{Z}_v(i,j)+\frac{1}{\xi}{\mathbf{N}_v(i,j)}|-\frac{\sigma}{\xi},0).
\end{align}\\
\textbf{Subproblem $\mathbf{E}_v$}: For given $\boldsymbol{\mathcal{G}}$, $\mathbf{Z}$, $\mathbf{M}$ , we can get:
\begin{align}\label{Ev}
\mathbf{E}^*=&\argmin_\mathbf{E}\lambda\|\mathbf{E}\|_{2,1}+\frac{\mu}{2}\|\mathbf{X}_v-\mathbf{X}_v\mathbf{Z}_v-\mathbf{E}_v\|_F^2 \notag\\
&+\sum_v{\langle\mathbf{Y}_v,\mathbf{X}_v-\mathbf{X}_v\mathbf{Z}_v-\mathbf{E}_v\rangle} \notag \\
=&\argmin_\mathbf{E}\lambda\|\mathbf{E}\|_{2,1}+\frac{1}{2}\|\mathbf{E}-\mathbf{D}\|_F^2,
\end{align}
where $\mathbf{D}$ is constructed by vertically concatenating the matrices $(\mathbf{X}_v-\mathbf{X}_v\mathbf{Z}_v+\frac{1}{\mu}\mathbf{Y}_v)$ .
This subproblem can be solved by \cite{LRRR}. \\
\begin{algorithm}[t]
\SetAlgoLined
\caption{\textbf{Multi-index fusion}}
\label{al-mmf}
\KwIn{Index matrix $\mathbf{X}_{v},~v=1,2,\ldots,V$, $\lambda > 0$, $\sigma>0$ , sparse threshold $threshold$, iters}
\KwOut{Fused index matrix $\mathbf{X}^*_{v},~v=1,2,\ldots,V$}
\BlankLine
\For {iter=1:iters}
{Initialized: $\mathbf{Z}_{v} = \mathbf{E}_{v} =\mathbf{Y}_{v} = \mathbf{M}_{v} =\mathbf{N}_{v} = \mathbf{0}$; $\boldsymbol{\mathcal{G}} = \boldsymbol{\mathcal{W}} = \mathbf{0}$;
 $\mu = 10^{-5}$, $\rho = 10^{-5}$,$\xi = 10^{-5}$, $\eta = 2$, $\mu_{\max} = \rho_{\max} =\xi_{\max}= 10^{10}$, $\varepsilon = 10^{-7}$;\\
\While{not converge}
{

        Update $\mathbf{Z}_{v},~v=1,2,\ldots,V$ by using (\ref{ZV});\\

    Update $\mathbf{E}$ by solving (\ref{Ev});\\


        Update $\mathbf{M}_{v},~v=1,2,\ldots,V$ by using (\ref{mv});

    Obtain $\boldsymbol{\mathcal{Z}} = \Phi(\mathbf{Z}_{1}, \mathbf{Z}_{2}, \ldots, \mathbf{Z}_{V})$\;
    Update $\boldsymbol{\mathcal{G}}$ via subproblem (\ref{tG})\;
    Update Lagrange multipliers $\boldsymbol{\mathcal{W}},\boldsymbol{\mathbf{Y}}_v,\boldsymbol{\mathbf{N}}_v,$$~v=1,2,\cdots,V$ by using (\ref{Ww})\;
    Update parameters $\mu$, $\xi$ and $\rho$: $\mu = \min(\eta \mu, \mu_{\max})$, $\rho = \min(\eta \rho, \rho_{\max})$, $\xi = \min(\eta \xi, \xi_{\max})$\;

    $(\mathbf{Z}_{1}, \ldots, \mathbf{Z}_{V}) = \Phi^{-1}(\boldsymbol{\mathcal{Z}})$,\ $(\mathbf{G}_{1}, \ldots, \mathbf{G}_{V}) = \Phi^{-1}(\boldsymbol{\mathcal{G}})$\;
    Check the convergence conditions:\\
\qquad $||\mathbf{X}_{v} - \mathbf{X}_{v}\mathbf{Z}_{v} - \mathbf{E}_{v}||_{\infty} < \varepsilon$ \\
\qquad $||\mathbf{Z}_{v} - \mathbf{G}_{v}||_{\infty} < \varepsilon$\;
\qquad $||\mathbf{Z}_{v} - \mathbf{M}_{v}||_{\infty} < \varepsilon$\;
}
$\mathbf{Z}_v^*=sparse(\mathbf{Z}_v),~v=1,2,\ldots,V$
$\mathbf{X}_i={\mathbf{X}}_i \sum_{v=1}^V(\mathbf{Z}_v^*+\mathbf{Z}_v^{*}{T}),~i=1,2,\ldots,V$\\
$\lambda=10\cdot\lambda$,
$\sigma=10\cdot\sigma$\\

}
$\mathbf{X}_v^*=sparse(\mathbf{X}_v),~v=1,2,\ldots,V$\\
\textbf{Return} Fused index matrix $\mathbf{X}^*_{v},~v=1,2,\ldots,V$.
\end{algorithm}
\textbf{Subproblem $\boldsymbol{\mathcal{G}}$}: At last, when the $\mathbf{E}$, $\mathbf{Z}$, $\mathbf{M}$ are fixed, we will solve the following subproblem for updating the tensor $\boldsymbol{\mathcal{G}}$,
\begin{equation}\label{tg}
\boldsymbol{\mathcal{G}}^*=\argmin_{\boldsymbol{\mathcal{G}}}\|\boldsymbol{\mathcal{G}}\|_{\circledast}+\frac{\rho}{2}\|\boldsymbol{\mathcal{G}}-(\boldsymbol{\mathcal{Z}}+\frac{1}{\rho}\boldsymbol{\mathcal{W}})\|.
\end{equation}
When we transform the Eq. (\ref{tg}) to the Fourier domain, it can be reformulated as:
\begin{align}\label{tG}
\boldsymbol{\mathcal{G}}_{f}^* = \argmin_{\boldsymbol{\mathcal{G}}_{f}} ~ \sum_{j=1}^{N}\tau'||\boldsymbol{\mathcal{G}}_{f}^{(j)}||_{*} + \frac{\rho}{2}||\boldsymbol{\mathcal{G}}_{f}^{(j)} - (\boldsymbol{\mathcal{Z}}+\boldsymbol{\frac{1}{\rho}\mathcal{W}})_{f}^{(j)}||_{F}^{2}.
\end{align}
Thus the tensor optimization can be divided into $N$ independent matrix subproblems in Fourier domain to solve. The procedure in \cite{mulview} can be applied to solve this subproblem.

In addition, the Lagrange multipliers are also need to be updated as:
\begin{align}
    \mathbf{Y}_{v}^{\ast} &= \mathbf{Y}_{v} + \mu(\mathbf{X}_{v} - \mathbf{X}_{v}\mathbf{Z}_{v} - \mathbf{E}_{v}), \notag\\
    \boldsymbol{\mathcal{W}}^{\ast} &= \boldsymbol{\mathcal{W}} + \rho(\boldsymbol{\mathcal{Z}} - \boldsymbol{\mathcal{G}}), \label{Ww}\\
    \mathbf{N}_{v}^{\ast} &= \mathbf{N}_{v} + \xi(\mathbf{Z}_{v} - \mathbf{N}_{v}). \notag
\end{align}

The above four steps are repeated until the convergence condition is satisfied. Although it is not easy to prove the convergence of the algorithm theoretically, two sufficient conditions suggested in \cite{LRRR} for our algorithm to converge are easily to be met fortunately.

Finally, the small value of functional matrix may not affect retrieval accuracy but will introduce the noise into the new index. So we simply set the value which is below a certain threshold $\theta_1$ to 0.
\begin{align}
sparse(\mathbf{Z}) = \left\{ {\begin{array}{*{20}{c}}
{0,{\rm{    }}\left| {{z_{ij}}} \right| < \theta_1}\\
{{z_{ij}},{\rm{  }}\left| {{z_{ij}}} \right| \ge \theta_1}
\end{array}} \right.
\end{align}
\subsection{Index Updating and Online Query}\label{final}
In the fusion process, 
we update our index matrix as follows,
\begin{equation}
\mathbf{X}^*_v=\mathbf{X}_v(\mathbf{I}+\frac{1}{V}\sum_v(\mathbf{Z}_v+\mathbf{Z}^T_v)),v=1,2,\ldots\,V.
\end{equation}
We iteratively fuse indexes for T times until we obtain the best retrieval accuracy. In each iteration, we execute normalization on each new index and expand 
the parameter $\lambda$ and $\sigma$ tenfold to guarantee the original representation of database images. When the fusion is finished, 
only one index is selected to be the final index for online retrieval. To leverage the inverted index structure and reduce the cost in memory and computation, we also set the elements of the final index below the threshold $\theta_2$ to zero.

Given the query image $q$, in the online query stage, we first extract only one type of visual feature $\mathbf{x}_v(q)$ used in our feature index. Then, we compute the consine similarity between the query and each database image.
It is worth noting that we can make full use of the high sparseness of feature indexes, {\it i.e.}, hypercolomn index. The computational complexity of calculating distance will be greatly reduced. At last, we sort the similarity scores in descending order and return the retrieval result. The entire fusion process is summarized in Algorithm 1.

\section{Experiment}\label{Experiment}
In this section, we perform experiments to present a comprehensive evaluation of the proposed method. Two applications ({\it i.e.} image retrieval and person re-identification) are tested, where we regard the person re-identification as a special retrieval task. The retrieval accuracy and memory consumption are evaluated for our approach in the retrieval benchmark datasets, while only search accuracy is tested on the Market-1501. Comparison is made to measure performance improvement to the baseline methods and some other state-of-the-art methods.
All experiments are implemented in Matlab on a workstation with Intel Xeon E5-2630 @ 2.30 GHz CPU, 128GB RAM, and TITANX GPU (12GB caches). To promote the culture of reproducible research, source codes and experimental results accompanying this paper will be released at https://www.researchgate.net/profile/Zhizhong\_Zhang5.

\subsection{Experimental Setup}
We evaluate the proposed algorithm on three public benchmark datasets {\it i.e.}, UKBench \cite{ukbench}, Holidays \cite{haming} and Market-1501 \cite{market}, where Market-1501 is one of the biggest person re-identification dataset. The UKBench dataset contains 10,200 images. All images in UKBench are taken as query and each of which has 4 relevant images.  We evaluate the retrieval accuracy by N-S score, which is the average number of relevant images of top 4 returned images. The Holidays dataset consists of 1,491 images taken from personal holidays photos, where 500 images are selected to be queries. Mean average precision (mAP) is adopted to evaluate the retrieval accuracy. The Market-1501 dataset is collected in front of a supermarket in Tsinghua University. Overall, this dataset contains 32,668 annotated bounding boxes of 1,501 identities. There are 12,936 images used for training and other 19732 images for testing. 
Both rank-1 error and mAP are adopted for evaluation. It is worth noting that we only use the testing images for training our multi-index fusion method.

\begin{table}[!ht]\label{finaltable}
\footnotesize
\renewcommand{\arraystretch}{1.2}
\begin{centering}
\begin{threeparttable}[]
\caption{Comparison of retrieval accuracy and memory cost. The performance of the comparison methods are taken from those original papers. The average SIFT features per image of comparison method is assumed to be 2,000. OQMC means online query memory cost for each indexed image. MMF means the index after our multi-index fusion. * means the baseline method}\label{dataset}
\tabcolsep=8pt
\begin{minipage}{0cm}
\begin{tabular}{@{}lccc}
\toprule[2pt] 
Methods       &UKBench(NS-score) &Holiday(mAP) &OQMC \\\hline
SIFT Index*  &3.03 &31.8\% &21.5KB \\
FC Index*   &3.42 &70.4\% &5.1KB \\
HC Index*    &3.28 &74.3\% &1.5KB \\ \hline
c-MI \cite{coupled}      &3.85 &85.8\%      &13.5KB \\
QSF \cite{rank}          &3.77 &84.6\%     &20KB\\
QaLF \cite{queryfusion}  &3.84 &88.0\%     &62KB  \\
CIE  \cite{calembed}     &3.86   &89.2\%    &4KB   \\
MFSMP   \cite{mulllindex}  &3.78 &78.8\%         &0.38KB\\
CoInd \cite{coindexing}    &3.60 &80.9\%      &24KB\\ \hline
\textbf{MMF-SIFT}  &\textbf{3.94} &84.8\% &10.1KB \\
\textbf{MMF-FC}   &3.92 &93.6\% &2.8KB \\
\textbf{MMF-HC}    &3.87 &\textbf{94.1}\% &1.2KB \\ \hline
\bottomrule[1pt]
\end{tabular}
\end{minipage}
\end{threeparttable}
\end{centering}
\end{table}

\subsection{Implementation Details}\label{detail}
In this section, we introduce some experiments detail such as the index building methods. On the UKBench and Holidays datasets, we extract three types of features to build our indexes separately. Specially, for \textbf{SIFT index}, we first extract the SIFT features \cite{hessin,sift}
and transform each SIFT descriptor with root-SIFT \cite{rootsift}. To avoid the loss in quantization, we assign each descriptor to three nearby visual words \cite{soft} with a pre-trained codebook\cite{akm}. Following the traditional Bag-of-Words method \cite{videogoogle}, we represent each database image as a 20K sparse vector in a TF-IDF manner \cite{videogoogle}. 
For CNN fully-connected index (\textbf{FC index}), we first resize each image to $224\times224$ and then pass it through the deep convolutional network, {\it i.e.}, AlexNet \cite{alex}, which is pre-trained on ImageNet by Caffe implementation \cite{caffe}. The outputs of the fully connected layers (FC6) are extracted and thus each dimension of feature vector can be regarded as a visual word. 
For Hypercolumn index (\textbf{HC index}), we use the VggNet \cite{vgg}, which is also pre-trained on ImageNet by Caffe implementation, as our Hypercolumn feature extractor.
The feature maps of conv $5\_4$ layer are extracted, whose size is $14\times14\times512$.
We take the activations of all filters $\mathbf{f}_{h}(m)\in \mathcal{R}^{512},m=1,2,\ldots,196$  as our feature vector. Then similar to the strategy of standard vector quantization, we quantize each $\mathbf{f}_{h}(m)$ to three nearest visual words of a pre-trained 10K codebook via TF-IDF weighting. 
For Market-1501 dataset, we follow three baseline methods proposed by \cite{reidreview,market}.
\renewcommand{\arraystretch}{1.5}
\begin{table}[tp]
  \centering
  \fontsize{8}{8}\selectfont
  \caption{Comparison of retrieval accuracy On Mraket-1501. Rank-1 error and mAP are used for evaluation. * means the baseline method.}
  \label{market}
    \begin{tabular}{|c|c|c|c|c|}
    \hline
    \multirow{2}{*}{Method}&
    \multicolumn{2}{c|}{Single Query}&\multicolumn{2}{c|}{ Multiple Query}\cr\cline{2-5}
    &Rank-1&mAP&Rank-1&mAP\cr
    \hline
    \hline
    BOW*\cite{market}&35.84\%&14.75\%&44.36\% &19.41\% \cr
    CaffeNet*\cite{reidreview} &49.36\%&32.10\%&66.63\%&41.25\%\cr
    ResNet50*\cite{reidreview} &74.02\%&49.36\%&81.26\%&59.10\%\cr\hline \hline
    NULL\cite{NULL} &61.02\%&35.68\%&71.56\%&46.03\%\cr
    Reranking\cite{rerank}&\textbf{77.11}\%&\textbf{63.63}\%&-&- \cr
    LBA\cite{lookba} &73.87\%&47.89\%&81.29\%&56.98\%\cr
    Gate-SCNN\cite{Gate-CNN}&65.88\%&39.55\%&76.04\%&48.45\%\cr
    S-LSTM\cite{S-LSTM}&-&-&65.6\%&35.31\%\cr
    SCSP\cite{SCSP}&51.9\%&26.35\%&-&-\cr
    CIE\cite{calembed}&73.77\%&57.55\%&79.39\%&65.24\%\cr
    SSDAL\cite{SSDAL}&39.40\%&19.60\%&49.00\%&25.80\%\cr\hline\hline
    \bf{MMF-BOW}&55.73\% &37.64\%&63.81\%&44.37\%\cr
    \bf{MMF-CaffeNet}&69.63\% &53.93\%&76.93\%&61.79\%\cr
    \bf{MMF-ResNet50}&\textbf{77.11}\% &62.39\%&\textbf{82.51}\%&\textbf{69.58}\%\cr
    \hline
    \end{tabular}
\end{table}


\subsection{Experimental Results on Image Retrieval}

For each image in the UKbench and Holidays datasets, we extract the CNN feature and SIFT feature, then perform the feature transform as aforementioned in section \ref{detail}. As shown in Table \ref{dataset}, our approach significantly outperforms the baseline methods on both UKBench and Holidays datasets. On UKBench dataset, we get the N-S score of 3.94, 3.92 and 3.87, which achieves absolute gain of $30.3\%$, $14.6\%$ and $18.0\%$, respectively.
For Holidays dataset, we increase the mAP of the baseline from $31.8\%$ to $84.8\%$ for SIFT index, from $70.4\%$ to $93.6\%$  for FC index and from $74.3\%$ to $94.1\%$ for HC index. It indicates that our MMF method could capture the complementarity between the SIFT feature and CNN feature, and elevate the performance to a higher level. More importantly, our approach shows the robustness for degenerate visual representation such as the SIFT index on Holidays. It is also worth noting that the baseline method have a great impact on the fusion result such as the high-dimensional MMF-SIFT index outperforms the other indexes on UKbench, while MMF-HC index achieves the best performance on Holiday.
The reason for this phenomenon is that the two datasets vary greatly {\it i.e.,} Holidays includes a very large variety of scene types, UKbench is a set of images containing relatively simple objects, which causes the performance of SIFT baseline on UKbench is much better than it on Holidays.

Moreover, our approach significantly outperforms the state-of-the-art feature fusion methods \cite{queryfusion,rank,coupled} with less online memory consumption. Although the proposed multi-index fusion method \cite{mulllindex} also provides comparable online memory consumption, its search accuracy is much worse than ours. Meanwhile, to achieve a accurate search result, the proposed CIE \cite{calembed} method requires elaborate baseline method ({\it i.e.,} N-S score of $3.53$ and $3.33$), which also demonstrates the effectiveness of the proposed method.
\begin{figure}[htb]\label{iteration}
\setlength{\abovecaptionskip}{-0pt}
\setlength{\belowcaptionskip}{-0pt}
\renewcommand{\figurename}{Figure}
\centering
\subfigure[]{
\includegraphics[width=0.24\textwidth]{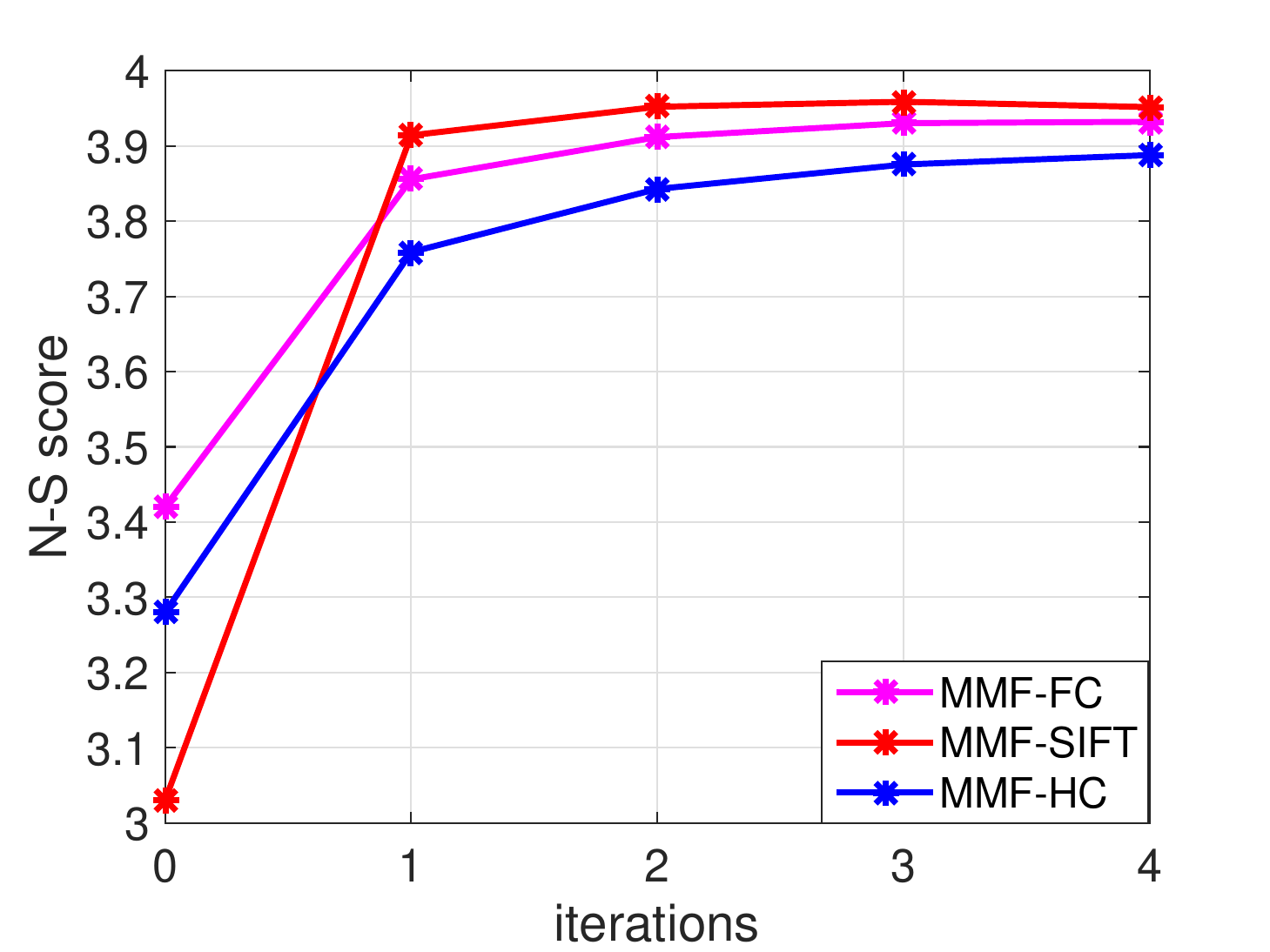}}
\hspace{-0.2in}
\subfigure[]{
\includegraphics[width=0.24\textwidth]{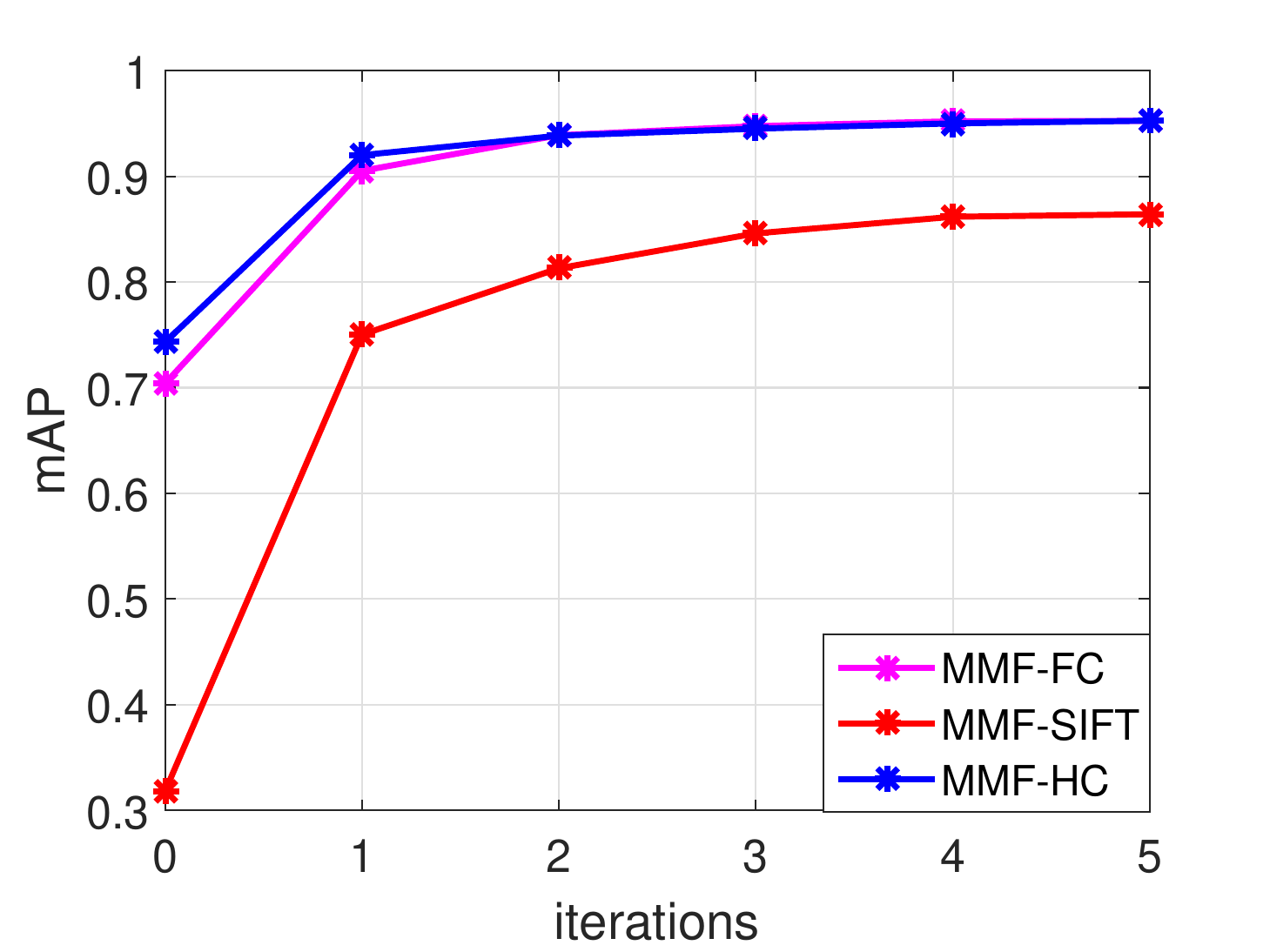}}

\label{iteration}
\subfigure[]{
\includegraphics[width=0.23\textwidth]{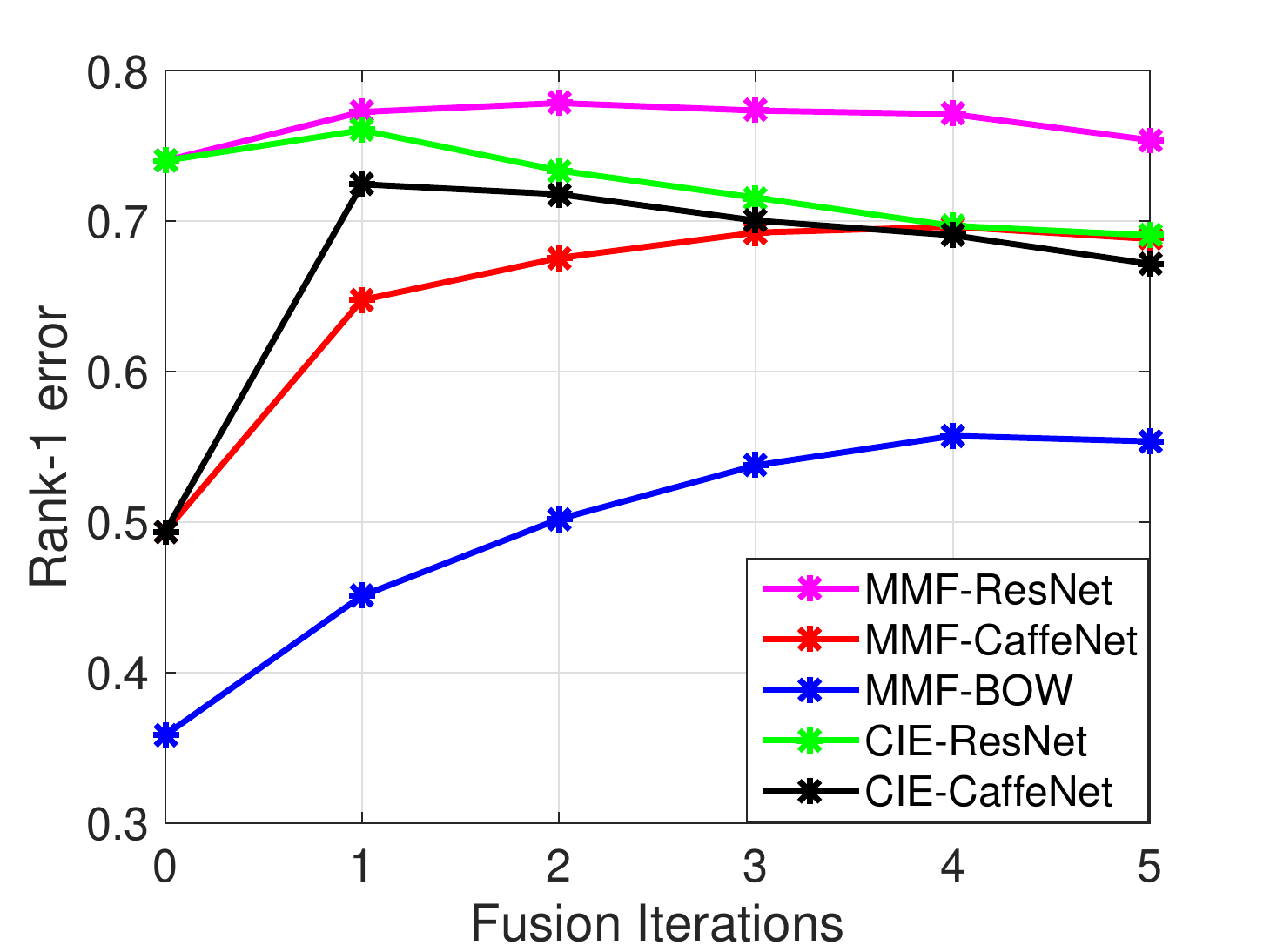}}
\hspace{-0.2in}
\subfigure[]{
\includegraphics[width=0.23\textwidth]{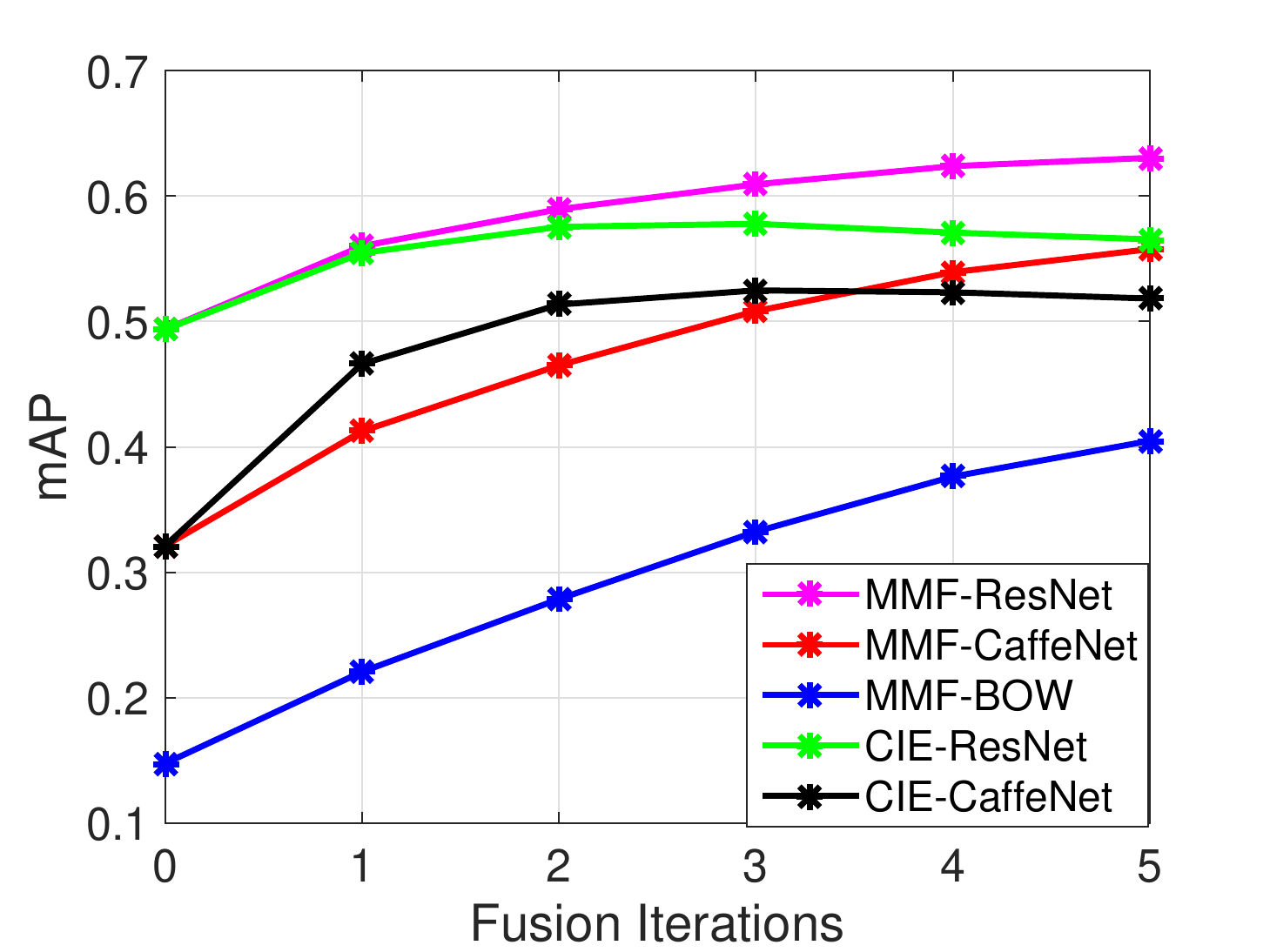}}
\caption{Influence of iteration number on retrieval accuracy on UKBench (a), Holiday (b) and Market-1501(c, d). CIE means the proposed Collaborative Index Embedding \cite{calembed}. Iteration 0 means the origin index. }
\label{iteration}
\end{figure}

In the online retrieval stage, the main memory cost of our method is to store the MMF index files. We assume each non-zero element of the feature vector in the index matrix takes 8 bytes to store the weight and image ID with the inverted index structure. After applying the sparse operation, our indexes require even less memory overhead than the origin index files, while keeping the competitive retrieval result. The online query computation complexity also gain the benefit from the sparsity of index files, which greatly reduce the query response time.

\subsection{Experimental Results on Person Re-identification}
\begin{figure*}[!htb]
  \setlength{\abovecaptionskip}{-5pt}
  \setlength{\belowcaptionskip}{-5pt}
  \centering
    \includegraphics[width=0.95\textwidth]{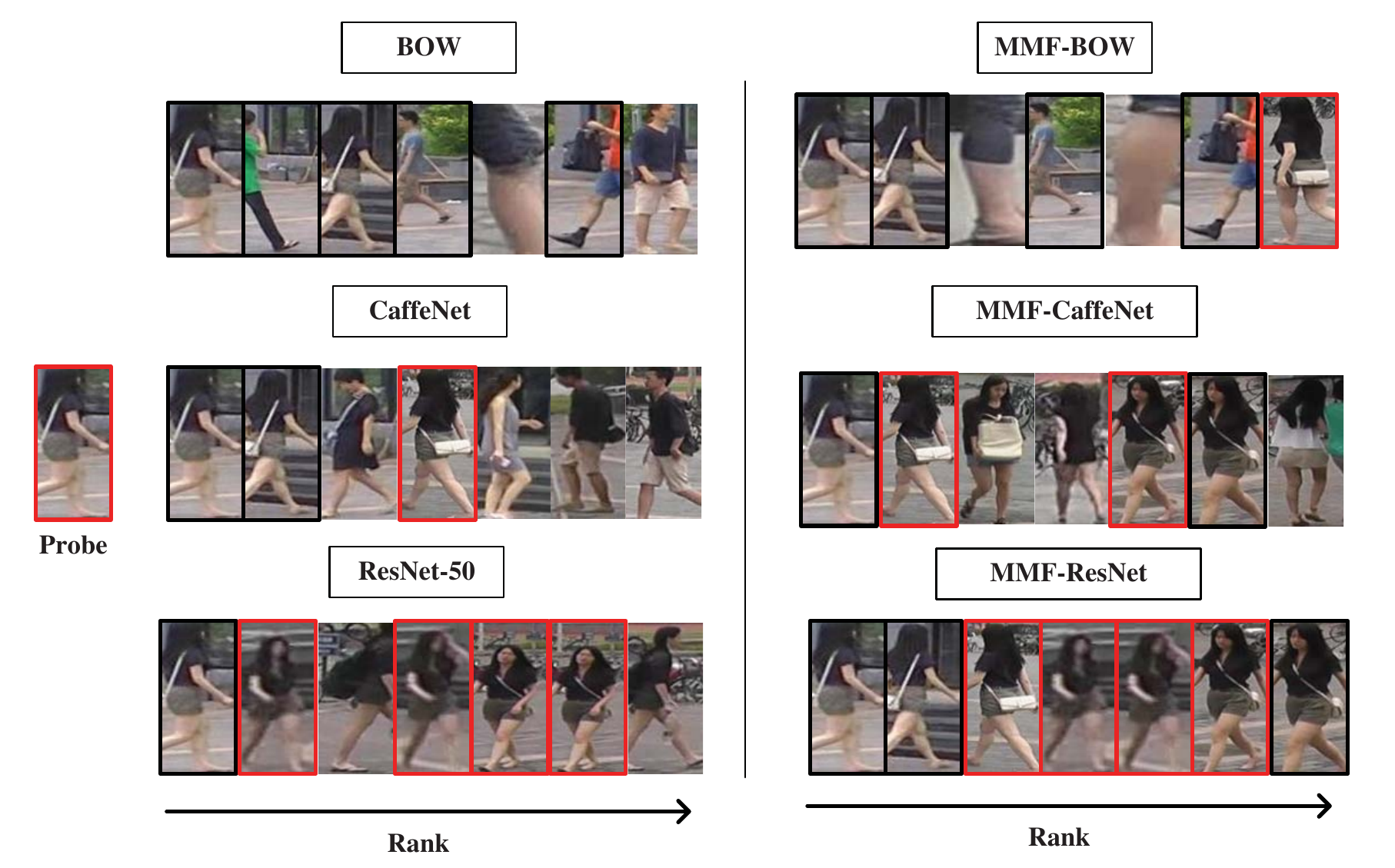}\\
  \caption{Representative retrieval results on the Market-1501 dataset. The black bounding box means the distractors or the images that come from the same camera with the query. The red bounding box means the true match persons and otherwise are wrong.} \label{reidresult} 
\end{figure*}

Following the protocol in \cite{reidreview,market}, we extracted three kinds of image features on Market-1501: the BOW feature, the ResNet50 feature and the AlexNet feature (CaffeNet feature). As shown in Table \ref{market}, the proposed method also outperforms baseline method on both single query and multiple query by a large margin. Although the CaffeNet and the BOW model obtain more improvement, the best Rank-1 error and mAP are still achieved by the MMF-ResNet50, which is also the best baseline method. However, the improvement of mAP is much higher than the improvement of Rank-1 error. While easy to understand, the proposed MMF can be regarded as an "off-line" query expansion or re-ranking technique, which can not fundamentally improve the discrimination of visual feature. The similar results are also presented by \cite{rerank}. From another perspective, if we can design more discriminative visual features, our multi-index fusion scheme can further improve its performance. Some representative retrieval results are shown in Fig. \ref{reidresult}, where the black bounding box means the distractors or the images that come from the same camera with the query, the red bounding box means the true match persons and otherwise are wrong. The complement information, such as the rank-1 and rank-2 images measured in CaffeNet, which includes the same person with the query, can be transferred through the different visual representations ({\it i.e.}, ResNet model).


As demonstrated in Table \ref{market}, the proposed approach achieves the comparable (even better) results with the state-of-the-art competitors, including the Gated Siamese Convolutional
Neural Network(Gate-SCNN) \cite{Gate-CNN}, Discriminative Null Space(NULL) \cite{NULL}, Spatially Constrained Similarity function on
Polynomial feature map(SCSP)\cite{SCSP}, Siamese Long Short-Term Memory (S-LSTM) \cite{S-LSTM}, Looking Beyond Appearances(LBA) \cite{lookba}, Deep attributes(SSDAL) \cite{SSDAL}, Re-ranking \cite{rerank} and the multi-index fusion method \cite{calembed}.
It is remarked that our method utilize the complementary information among different visual representation and perform at off-line stage, while the re-ranking \cite{rerank} method takes advantage of the probe information and must perform at on-line stage.

\subsection{Parameter Analysis}
In this section, we discuss the impact of parameters for our approach. Five key parameters influence the performance of the retrieval system, including the parameter $\lambda$, $\sigma$ in Eq. (\ref{ourmodel}), the threshold $\theta_1$ for functional matrix, the threshold $\theta_2$ for the final index and the fusing iteration number $T$ .

We first evaluate the impact of fusing iteration number $T$ for the search accuracy. As shown in Fig. \ref{iteration}, the retrieval accuracy on both UKbench and Holidays first quickly rises to the peak and then keep stable with the increase of fusing iterations. On UKBench, we fuse $3$ times until the MMF-SIFT index obtain the best performance but there is still room for HC index to improve. On Holidays, we fuse 4 times when the performance of all the indexes are keeping stable. As for the Market-1501 dataset, the Rank-1 error of MMF ResNet slightly drops after $T=2$, which is even worse in \cite{calembed}. But the mAP of all the indexes keep improving due to the characteristic of the multi-index fusion methods. We choose $T=4$ for Market-1501 for relatively stable performance.
\begin{figure}[htb]
\setlength{\abovecaptionskip}{-0pt}
\setlength{\belowcaptionskip}{-0pt}
\renewcommand{\figurename}{Figure}
\centering
\subfigure[]{
\includegraphics[width=0.24\textwidth]{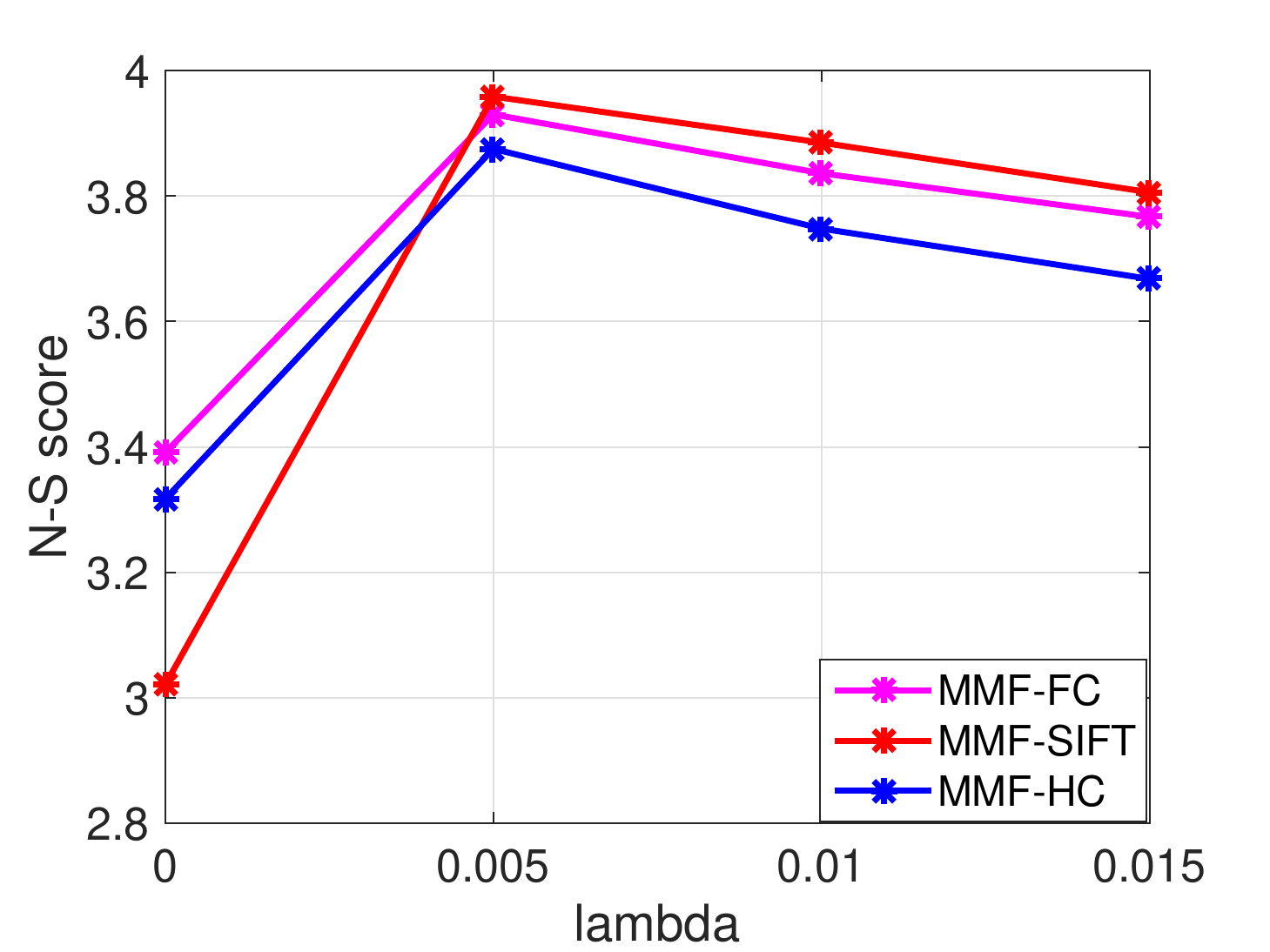}}
\hspace{-0.2in}
\subfigure[]{
\includegraphics[width=0.24\textwidth]{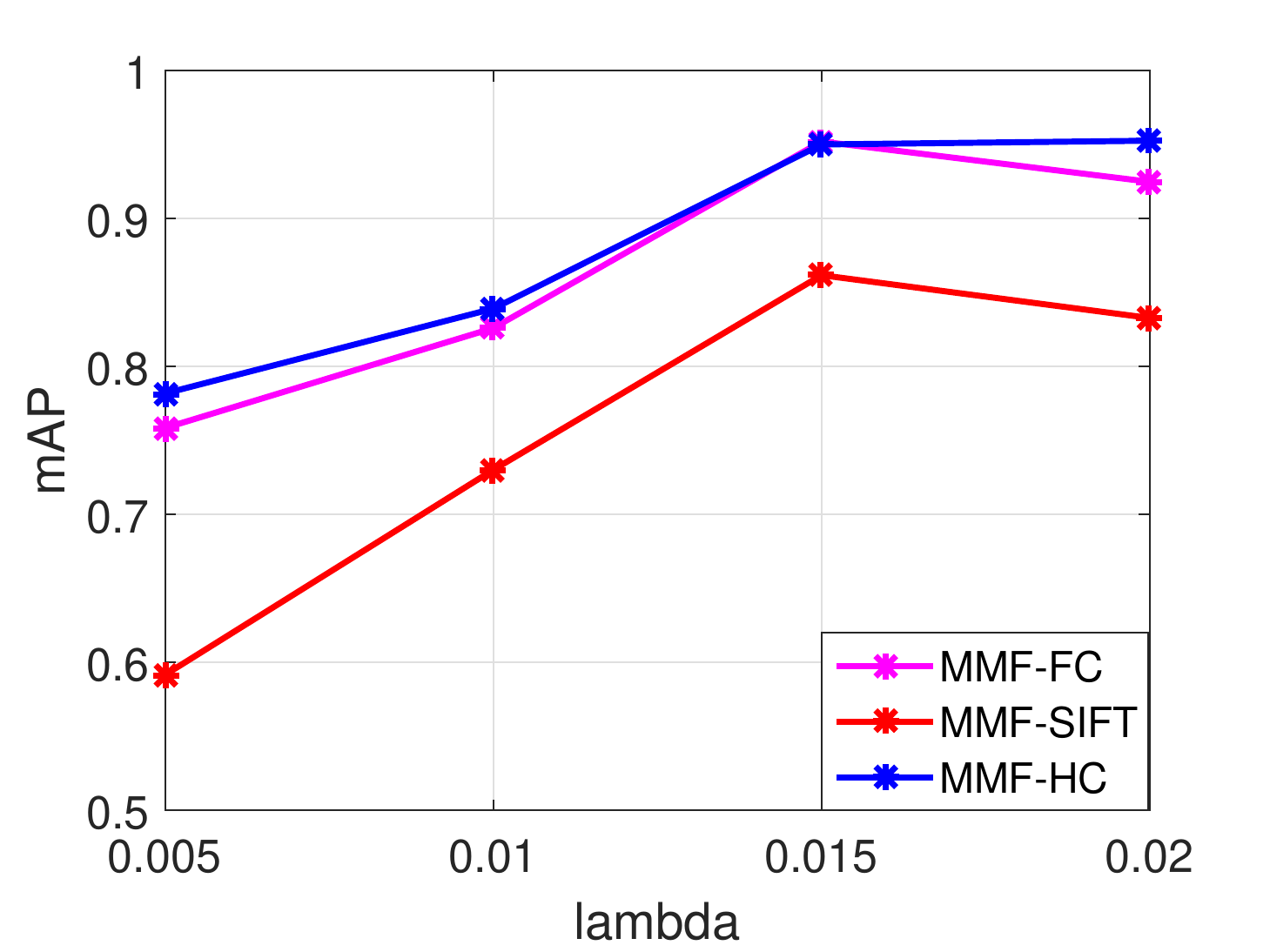}}
\subfigure[]{
\includegraphics[width=0.24\textwidth]{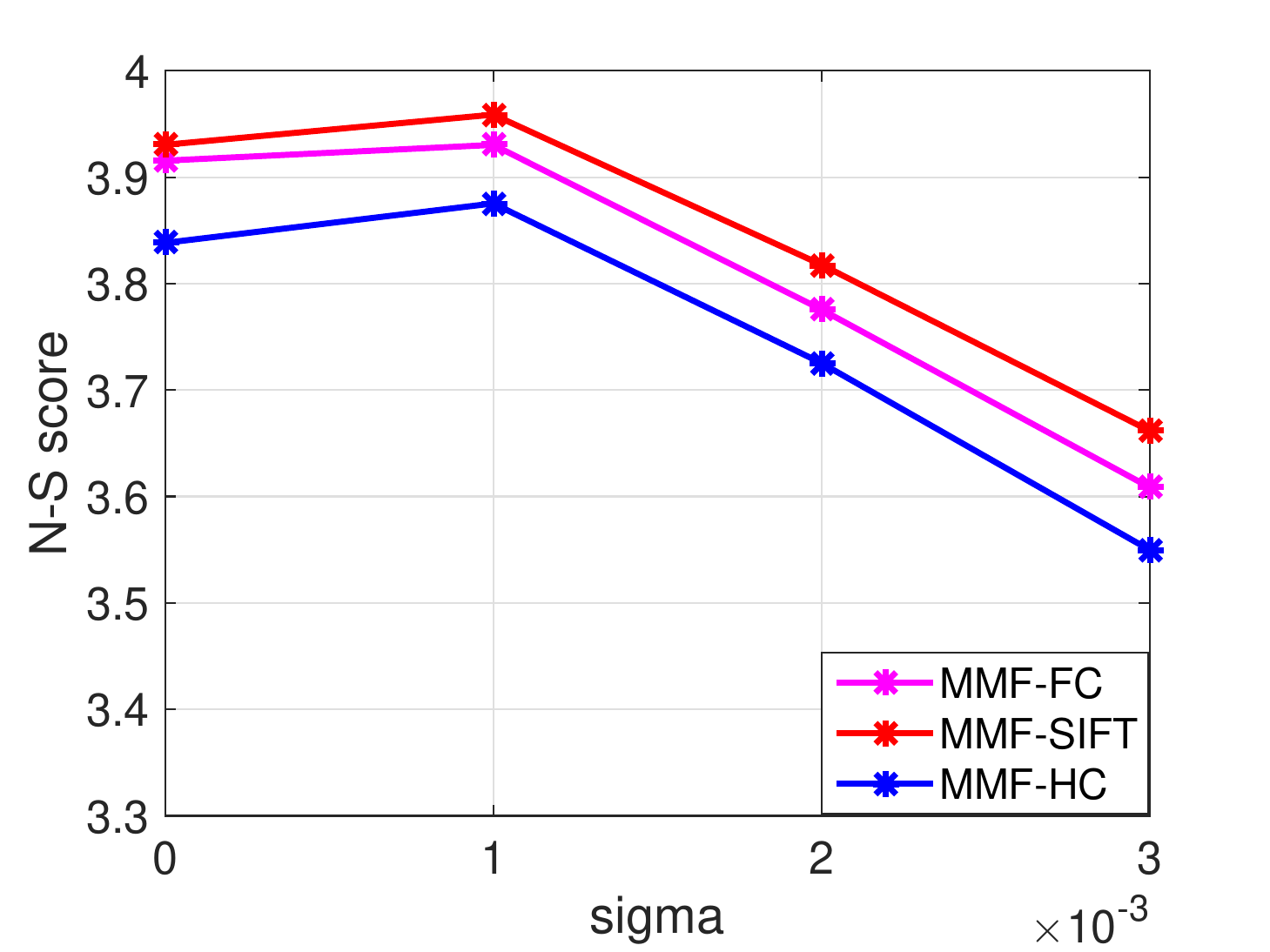}}
\hspace{-0.2in}
\subfigure[]{
\includegraphics[width=0.24\textwidth]{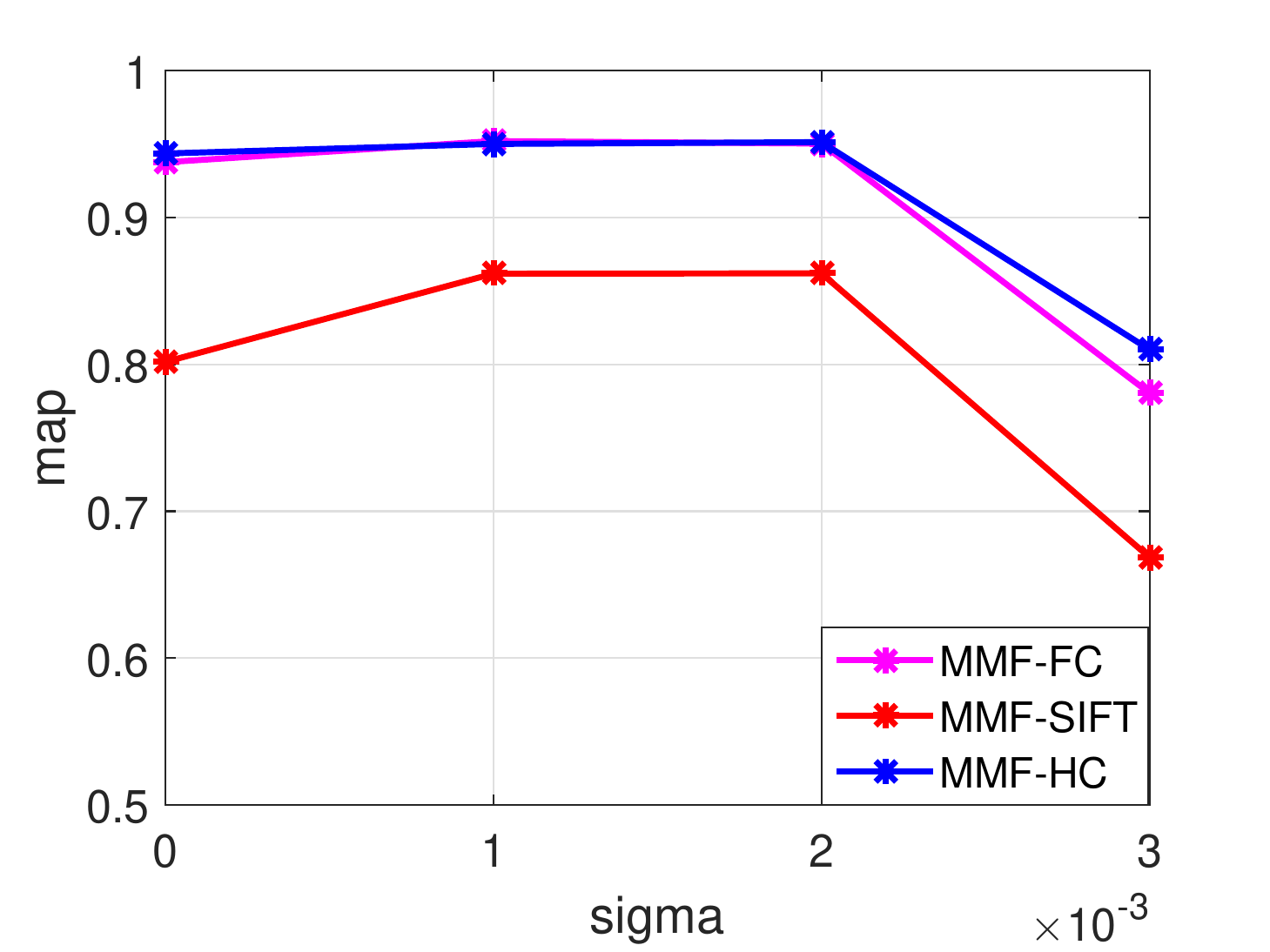}}

\caption{Influence of $\lambda$, $\sigma$ on retrieval accuracy on UKBench (left) and Holiday (right)}\label{lambda}
\end{figure}
\begin{figure}[htb]
\setlength{\abovecaptionskip}{0pt}
\setlength{\belowcaptionskip}{0pt}
\renewcommand{\figurename}{Figure}
\centering
\subfigure[]{
\includegraphics[width=0.24\textwidth]{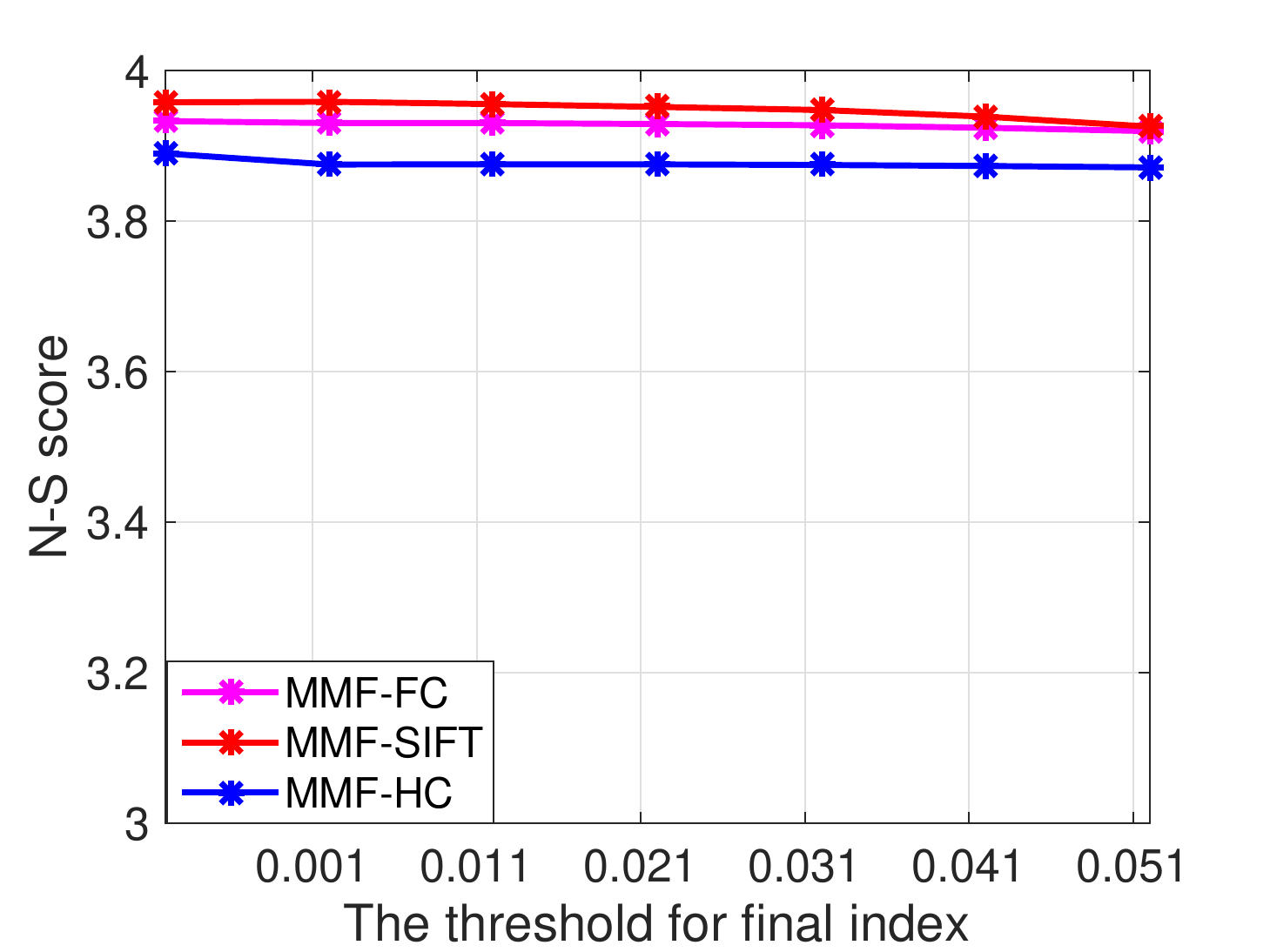}}
\hspace{-0.2in}
\subfigure[]{
\includegraphics[width=0.24\textwidth]{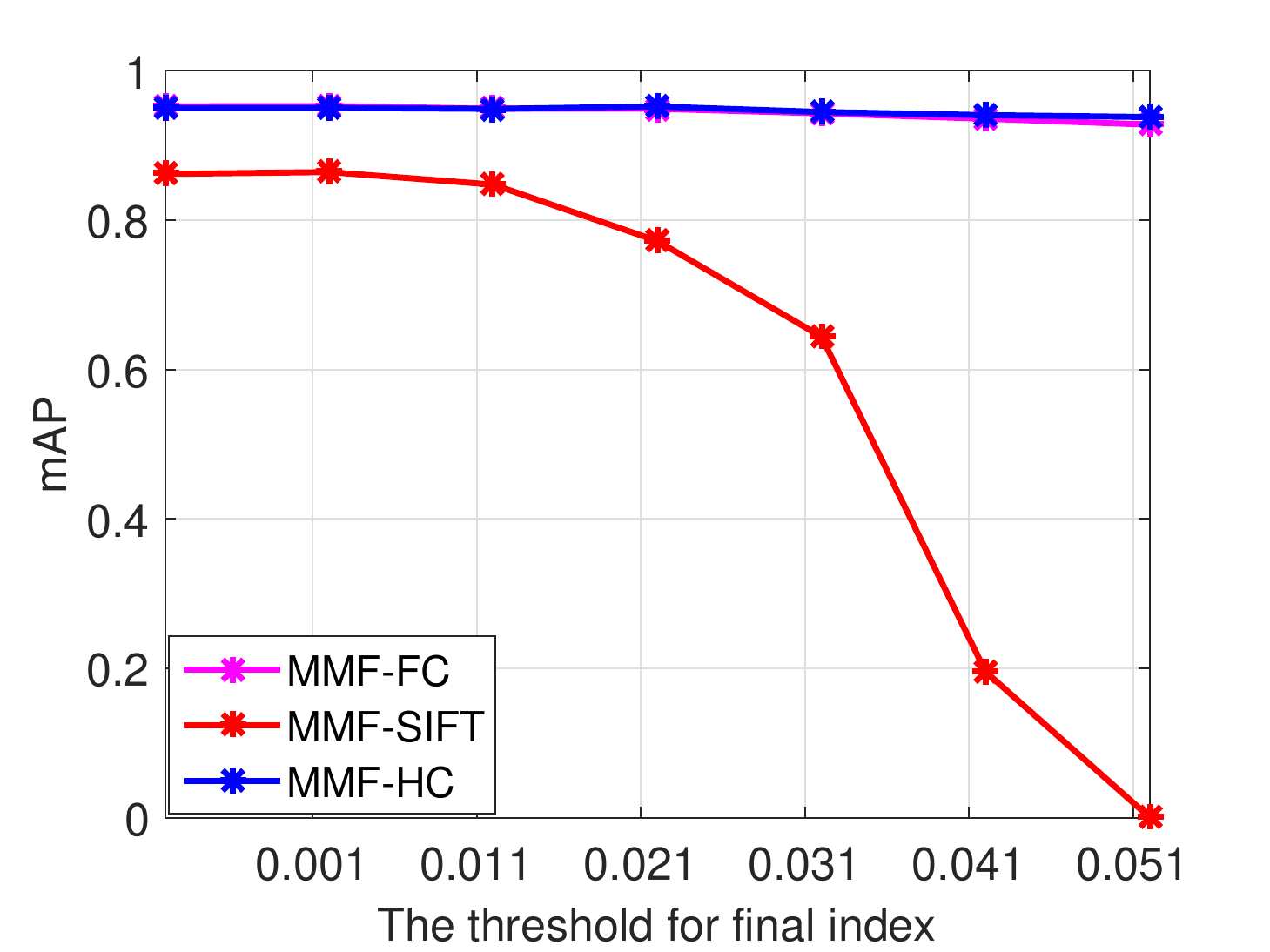}}
\subfigure[]{
\includegraphics[width=0.24\textwidth]{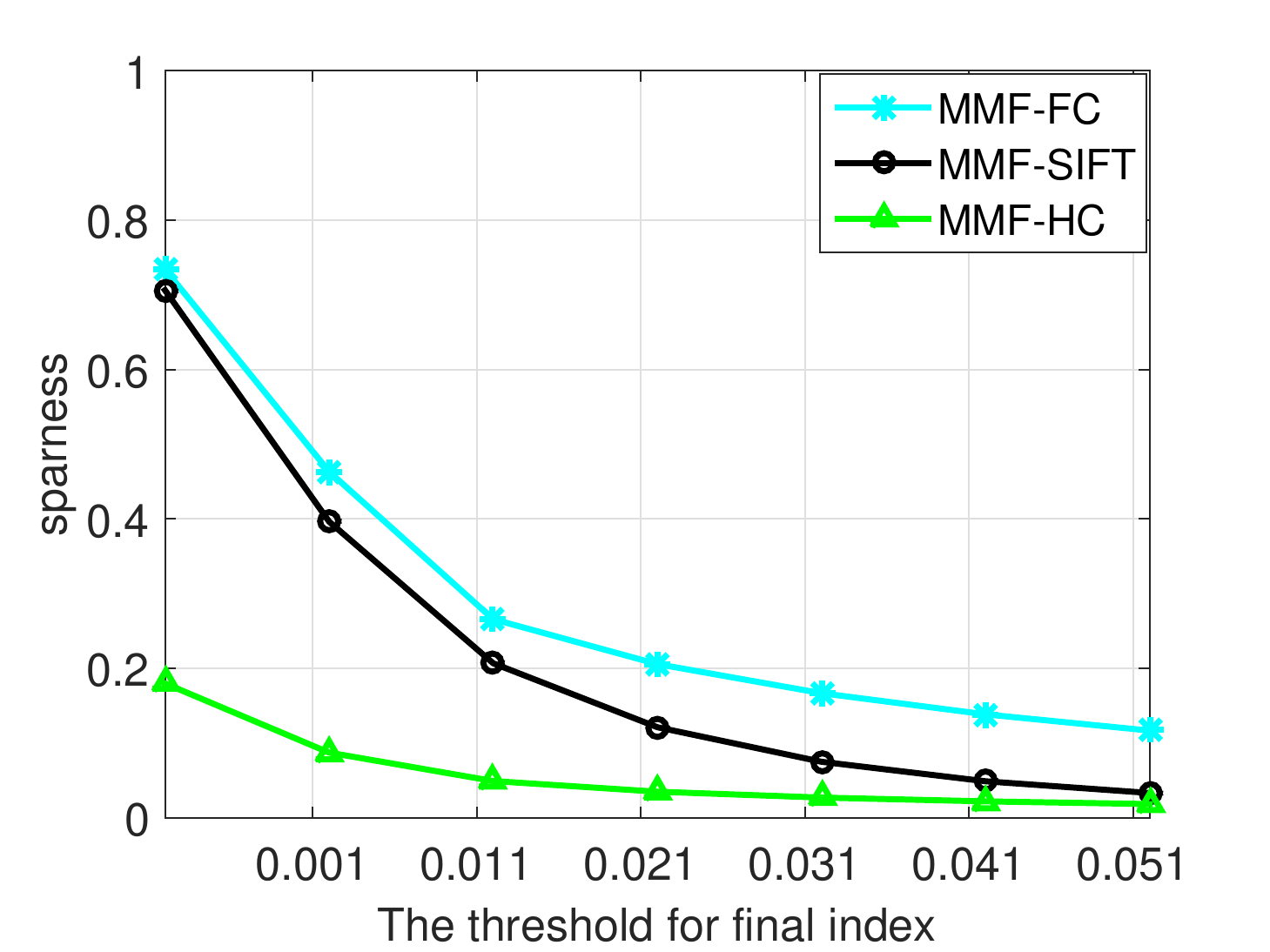}}
\hspace{-0.2in}
\subfigure[]{
\includegraphics[width=0.24\textwidth]{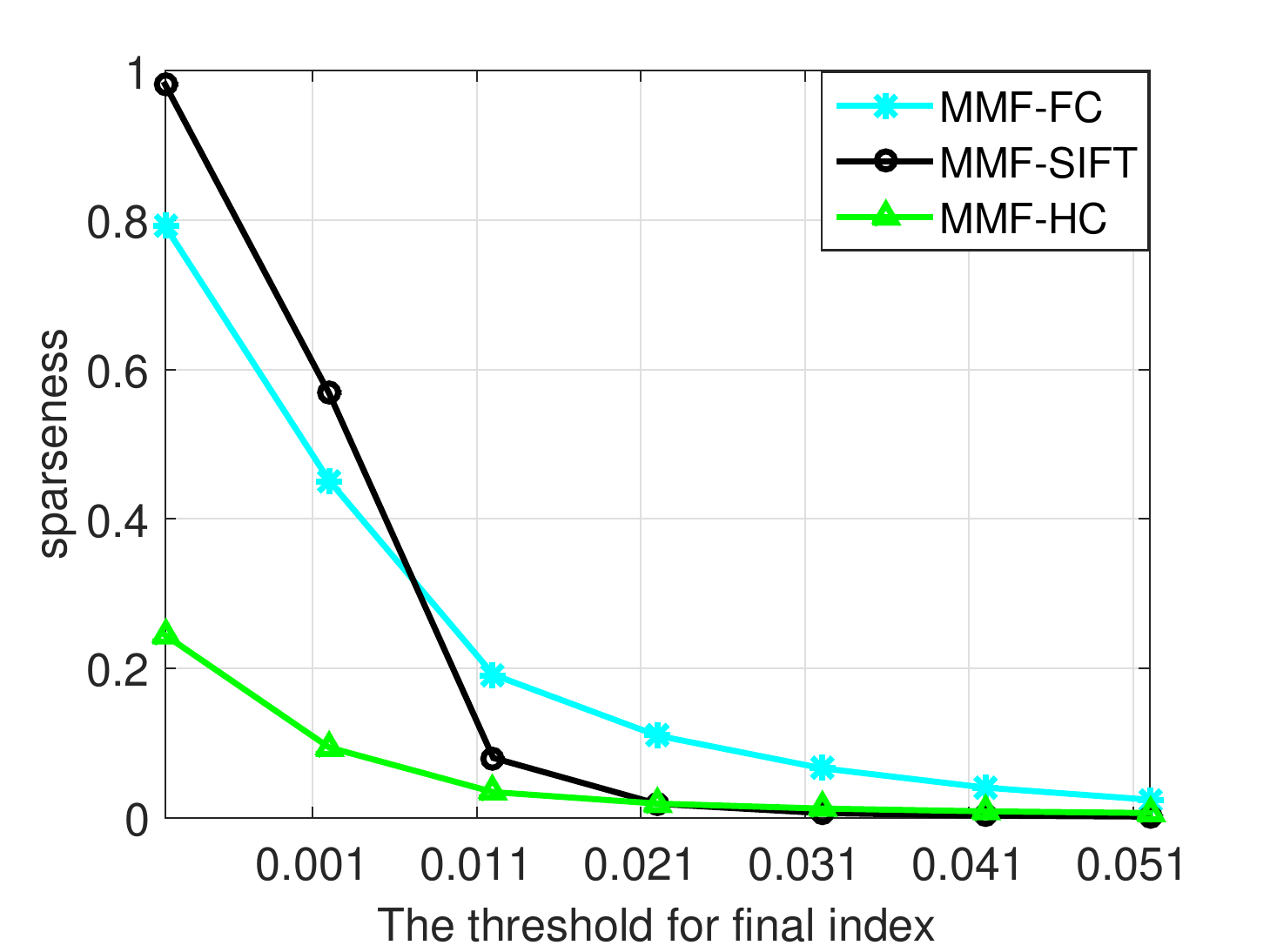}}

\caption{Influence of threshold for final index on the sparseness and retrieval accuracy on UKBench (left) and Holiday (right)}\label{sparse}
\end{figure}
Although the CIE method achieves comparable result, by setting the parameter $\alpha=\beta=0.4,~p=q=9,~m=20$ (with a simple grid search), its performance of Rank1-error drops sharply after a few iterations.

We then evaluate the impact of reconstruction error parameters $\lambda$ and sparse parameters $\sigma$ by using different values of $\lambda$ and $\sigma$. Although the parameter $\lambda$ and $\sigma$ play an important role on performance, most results are still much better than the baseline methods. As shown in Fig. \ref{lambda}, when the $\lambda$ increases, the retrieval accuracy firstly climbs to the peak point and then slowly decreases on both datesets by fixing the $\sigma=0.001$. The reason for this phenomenon is that the larger $\lambda$ is, the less modification is made to the index. As $\lambda$ decreases to $0$, the functional matrix is degenerated to identity matrix. The influence of parameter $\sigma$ shows similar performance as $\lambda$ on both datasets by fixing the $\lambda=0.005$ and 0.015 respectively. But when $\sigma$ increases to a certain extent, all the values of functional matrix have been suppressed due to excessive sparsity constraints, which cause the retrieval accuracy, {\it i.e.}, MMF-SIFT index drops sharply. For Market-1501 dataset, we set $\lambda=0.010$ and $\sigma=0.001$ to obtain the best performance. Empirically, $\lambda$ is often locate at $0.005$ to $0.015$  and $\sigma=0.001$ is suitable for most situations.

Fig. \ref{sparse} shows the influence of threshold $\theta_2$ for final index on the sparseness and retrieval accuracy of index matrix. We can easily get that the larger fusing iteration number, the less sparseness of the index structure, where this phenomenon is also presented in \cite{mulllindex,calembed}.
When the threshold increases, it is observed that the sparseness of indexes drop sharply while the retrieval accuracy keeps stable on both two datasets. The situation on Holidays is slightly different from it on UKBench, especially for the MMF-SIFT index due to the fact that the larger fusing iteration number and the larger SIFT descriptors number, which smooth the energy. The influence of threshold $\theta_1$ for functional matrix will keep stable with a little performance boosting when it varies from 0.005 to 0.02. The detail analyse will be presented in section \ref{Discussion}.

\begin{figure}[!htb]
  \setlength{\abovecaptionskip}{-0pt}
  \setlength{\belowcaptionskip}{-0pt}
  \centering
    \includegraphics[width=0.45\textwidth]{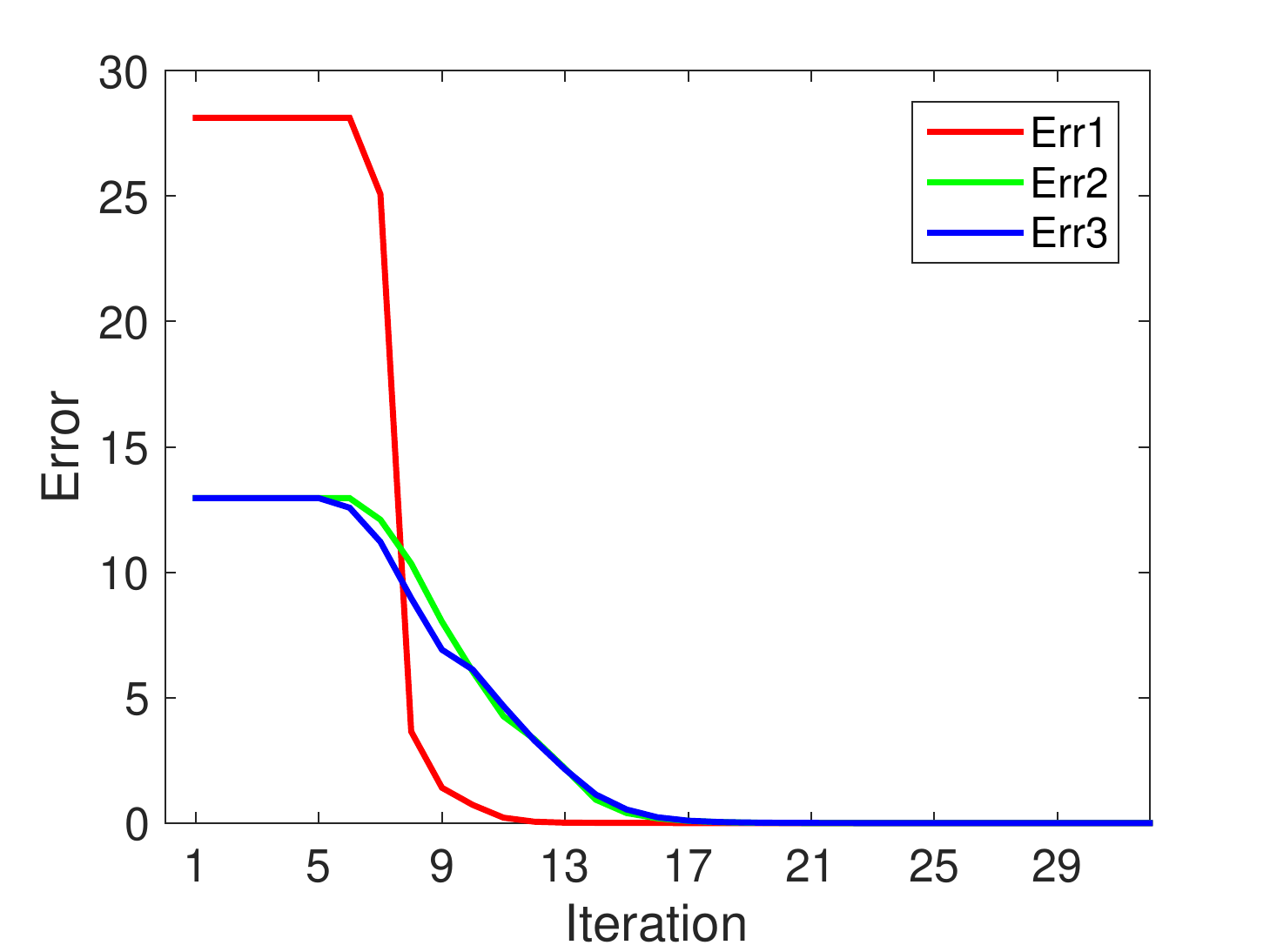}\\
  \caption{Convergence curves on Holiday dataset.} \label{convergencess} 
\end{figure}

In addition, our optimization method converges fast by setting the parameter as Algorithm \ref{al-mmf} states, which is illustrated in Fig. \ref{convergencess}. Three curves record the errors (defined in Eq. (\ref{convergences})) in each iteration step.
\begin{align}\label{convergences}
&\text{Err}_1=||\mathbf{X}_{v} - \mathbf{X}_{v}\mathbf{Z}_{v} - \mathbf{E}_{v}||_{\infty}  \notag\\
&\text{Err}_2=||\mathbf{Z}_{v} - \mathbf{G}_{v}||_{\infty} \\
&\text{Err}_3=||\mathbf{Z}_{v} - \mathbf{M}_{v}||_{\infty}  \notag
\end{align}
\begin{figure}[!htb]
  \setlength{\abovecaptionskip}{0pt}
  \setlength{\belowcaptionskip}{-0pt}
  \centering
    \includegraphics[width=0.48\textwidth]{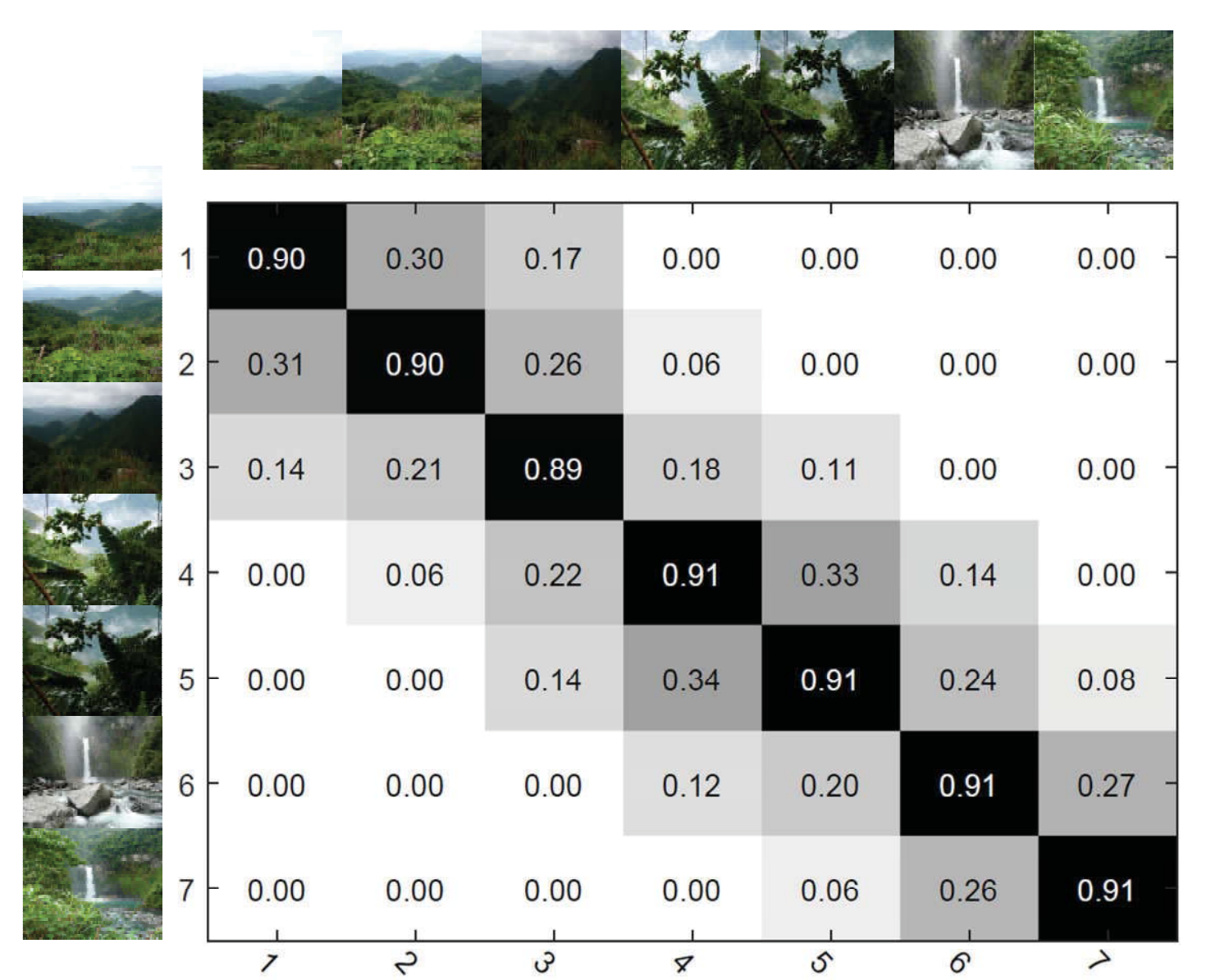}\\
  \caption{Representative functional matrix $\mathbf{Z}$ learned on the Holidays datasets. Larger values indicate two images are more positively correlated. } \label{functionalmatrix} 
\end{figure}

\subsection{Discussion and Analyses}\label{Discussion}
We conduct further analyses and experiments to better understand the characteristics of our Multi-index fusion scheme.\\

\textbf{Scalability}: Although the updating scheme seems costly, as discussed above, the whole procedure only perform once at off-line training time. Meanwhile, in Algorithm \ref{al-mmf}, the inverse matrix can only be calculated once during the whole iteration with proper parameter. The most time consuming part of our method is to solve the subproblem $\boldsymbol{\mathcal{Z}}$, but it equals to calculate $\frac{(N-1)}{2}$ matrix SVD, whose dimension is $N \times V$. This special structure can be easily parallelized and will be invested in our future work. In summary, it takes $\mathcal{O}(2N^2 V\log(N)) $ for calculating the FFT and its inverse. Take $\mathcal{O}(N^2 V^2)$ for calculating the matrix SVD.
As for the subproblem $\mathbf{E}$ and $\mathbf{M}_v$, they take $\mathcal{O}(N^2 V)$ in each iteration. Since $\log(N)\gg V$, the complexity of our MMF method is:
 \begin{equation}
 \mathcal{O}(TK(2N^2 V\log(N))),
 \end{equation}
 \begin{figure}[!htb]
  \setlength{\abovecaptionskip}{0pt}
  \setlength{\belowcaptionskip}{-0pt}
  \centering
    \includegraphics[width=0.48\textwidth]{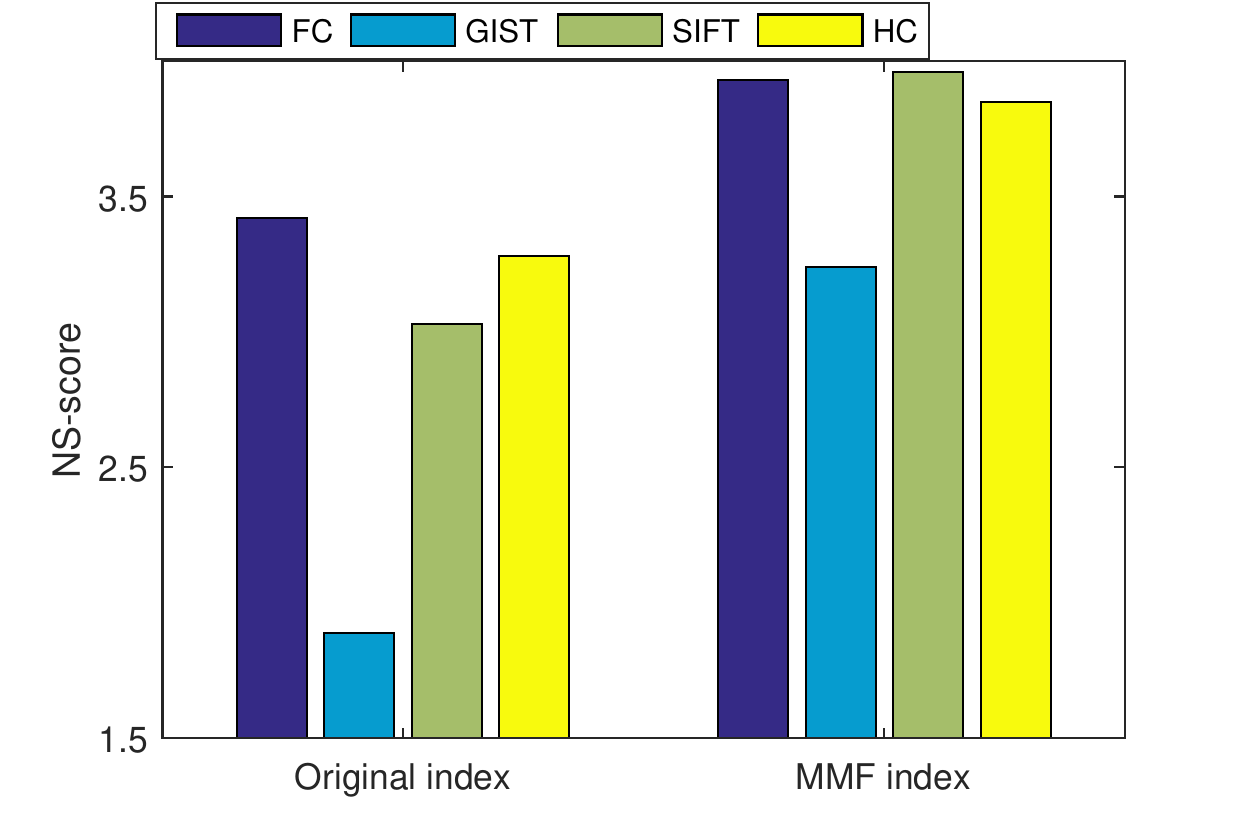}\\
  \caption{Comparison between the original index and the proposed MMF inedx in terms of accuracy on UKbench dataset. } \label{barz} 
\end{figure}
where $K$ means the iteration number. In practice, $T$ usually locates at 3-4 and $K$ locates at 30-50.
More importantly, as shown in Fig. \ref{functionalmatrix}, a block matrix structure is presented, which clearly demonstrates the sample-dependence assumption. Thus we can divide the dataset into image groups to further reduce the computation and memory consumption without incurring the performance lost. This will be invested in our future work.\\
\textbf{Robust}: We also extract GIST feature as the $4$th index for retrieval on UKbench and get a NS-score of 1.89. When we iteratively fuse three times, its performance improve to 3.24. While for the other indexes, we almost achieve the same result with only 0.01 absolute N-S score reduction as shown in Fig. \ref{barz}. This phenomenon demonstrates the robustness of our method and verify the index-specific dependence assumption implicitly.\\
\textbf{Analysis on functional matrix $Z$}: To evaluate the effect of sparsity of functional matrix in our model, we conduct experiments with variants of our approach. On Market-1501, we remove the sparse operation on the functional matrix, it drops a almost $5\%$ absolute reduction on Rank1-error. It indicates that the sparsity of functional matrix not only play an important role for scalability, but also the performance of the multi-index fusion framework. In practice, the sparsity of our learned functional matrix have been identified as shown in Fig. \ref{functionalmatrix}. Experiments demonstrate almost 97\% elements of our learned functional matrix are zero on Market-1501, which clearly reveals the subspace structure in the gallery. Furthermore, only a few relevant images have been updated, which may slightly destroy the original visual representation.\\
\textbf{Limitations and Future work}: Although the proposed method achieves impressive performances, there are still some limitations in this work, which will be further invested in the future work. First of all, without the ground-truth, it is hardly to choose the final index, especially when we meet the poor baseline method ({\it e.g.,} GIST index). Second, the sparseness of the visual index will not be guaranteed directly during the iteration. The sparse operation will destroy the original image representation and compel the related  images to be the same. For these limitations, we can use the priori knowledge and the larger $\theta_1$ to solve, but need further investigation. 

\section{Conclusion}\label{conclusion}
In this paper, a t-SVD based tensor optimization is proposed to tackle the multi-index fusion problem for image retrieval. 
Our proposed technique, MMF, inherits the core idea of CIE \cite{calembed}, that is, fuse different visual representations on index level. Furthermore, MMF explores the high-order information assumed by index-specific and sample-specific dependence to capture the complementary shared by different visual feature. Different from traditional multi-index fusion approach, the proposed method find an optimal functional matrix, which is optimized in a unified tensor space, to propagate similarities and update the indexes with sparse constraint.
Experimental results reveal that our approach significantly outperforms baseline methods and some other state-of-the-art methods in retrieval accuracy, and with little additional memory cost in online query stage.
Future research will include the following: 1) the parallel computing for t-SVD; 2) the final index selection method; 3) the strategy of splitting images into groups for scalable image retrieval.


%

\ifCLASSOPTIONcaptionsoff
  \newpage
\fi


\begin{thebibliography}{1}


\bibitem{hessin}
K. Mikolajczyk, C. Schmid. Scale affine \& invariant interest point detectors. \hskip 1em plus
  0.5em minus 0.4em\relax {\it International journal of computer vision}, vol. 60, no. 1, pp. 63-86, 2004.

\bibitem{fisher}
F. Perronnin, C. Dance. Fisher kernels on visual vocabularies for image categorization. {\it Proceedings of the IEEE Conference on Computer Vision and Pattern Recognition}, 2007.

\bibitem{deepr}
F. Radenovi$\acute{c}$, G. Tolias, O. Chum. CNN image retrieval learns from BoW: Unsupervised fine-tuning with hard examples. {\it European Conference on Computer Vision}, 2016.

\bibitem{VLAD}
H. J$\acute{e}$gou, F. Perronnin, M. Douze, et al. Aggregating local image descriptors into compact codes. {\it IEEE Transactions on Pattern Analysis and Machine Intelligence}, vol. 34, no. 9 pp. 1704-1716, 2012.

%
\bibitem{rank}
S. Zhang, M. Yang, T. Cour, K. Yu, D. N. Metaxas. Query specific rank fusion for image retrieval. {\it IEEE Transactions on Pattern Analysis and Machine Intelligence}, vol. 34, no. 9, pp. 803-815, 2015.

\bibitem{Mulfeature}
Y. Yang, J. Song, Z. Huang, Z. Ma, N. Sebe, A. G. Hauptmann. Multi-Feature Fusion via Hierarchical Regression for Multimedia Analysis. {\it IEEE Trans. on Multimeadia}, vol. 15, no. 3, pp. 572-581, 2012.

\bibitem{tembedding}
H. J$\acute{e}$gou, A. Zisserman. Triangulation embedding and democratic aggregation for image search. {\it Proceedings of the IEEE Conference on Computer Vision and Pattern Recognition}, 2014.

\bibitem{sift}
D. Lowe. Distinctive image features from scale-invariant keypoints. \hskip 1em plus
  0.5em minus 0.4em\relax {\it International journal of computer vision},  vol. 60, no. 2, pp. 91-110, 2004.
%
\bibitem{akm}
J. Philbin, O. Chum, M. Isard, J. Sivic, A. Zisserman. Object retrieval with large vocabularies and fast spatial matchin. \hskip 1em plus
  0.5em minus 0.4em\relax {\it Proceedings of the IEEE Conference on Computer Vision and Pattern Recognition}, 2007.

%
\bibitem{rootsift}
R. Arandjelovi$\acute{c}$, A. Zisserman. Three things everyone should know to improve object retrieval. {\it Proceedings of the IEEE Conference on Computer Vision and Pattern Recognition}, 2012.

\bibitem{queryfusion}
L. Zheng, S. Wang, L. Tian, F. He, Z. Liu, Q. Tian. Query-adaptive late fusion for image search and person re-identification. {\it Proceedings of the IEEE Conference on Computer Vision and Pattern Recognition}, 2015.

\bibitem{coupled}
L. Zheng, S. Wang, Z. Liu, Q. Tian. Packing and Padding: Coupled Multi-Index for Accurate Image Retrieval.  {\it Proceedings of the IEEE Conference on Computer Vision and Pattern Recognition}, 2014.

\bibitem{neuralcodes}
A. Babenko, A. Slesarev, A. Chigorin, V. Lempitsky. Neural codes for image retrieval. {\it European conference on computer vision}, 2014.

\bibitem{soft}
J. Philbin, O. Chum, M. Isard, J. Sivic, A. Zisserman. Lost in quantization: Improving particular object retrieval in large scale image databases View Document. {\it Proceedings of the IEEE Conference on Computer Vision and Pattern Recognition}, 2008.
%
\bibitem{videogoogle}
J. Sivic, A. Zisserman. Video Google: A text retrieval approach to object matching in videos. {\it Proceedings of the IEEE Conference on Computer Vision and Pattern Recognition}, 2003.

%
\bibitem{binary}
F. Shen, W. Liu, S. Zhang, Y. Yang, H. T. Shen. Learning binary codes for maximum inner product search. {\it Proceedings of the IEEE International Conference on Computer Vision
}, 2015.

\bibitem{alex}
A. Krizhevsky, I. Sutskever, G. E. Hinton. Imagenet classification with deep convolutional neural networks. {\it Advances in neural information processing systems}, 2012.

%
\bibitem{vgg}
K. Simonyan, A. Zisserman. Very deep convolutional networks for large-scale image recognition. {\it arXiv preprint arXiv:1409.1556}, 2014.

%


%
\bibitem{caffe}
Y. Jia, E. Shelhamer, J. Donahue, S. Karayev, J. Long, R. Girshick, S. Guadarrama, T. Darrell. Caffe: Convolutional architecture for fast feature embedding. {\it Proceedings of ACM international conference on Multimedia}, 2014.
%
\bibitem{local}
J. Ng, F. Yang, L. Davis. Exploiting local features from deep networks for image retrieval. {\it Proceedings of the IEEE Conference on Computer Vision and Pattern Recognition}, 2015.

\bibitem{ddaig}
Z. Gao, J. Xue, W. Zhou, S. Pang, Q. Tian. Democratic diffusion aggregation for image retrieval. {\it IEEE Trans. on Multimedia}, vol. 18, no. 8, pp. 1661-1674, 2016.

\bibitem{LRRR}
G. Liu, Z. Lin, S. Yan, J. Sun, Y. Yu, Y. Ma. Robust recovery of subspace structures by low-rank representation. {\it IEEE Trans. on Pattern Analysis and Machine Intelligence}, vol 35, no. 1, pp. 171-184, 2013.
%
\bibitem{mulview}
Y. Xie, D. Tao, W. Zhang, L. Zhang. On Unifying Multi-View Self-Representations for Clustering by Tensor Multi-Rank Minimization. {\it arXiv preprint arXiv:1610.07126}, 2016.

%
\bibitem{tsvd}
M. E. Kilmer and C. D. Martin. Factorization strategies for third-ordertensors. {\it Linear Algebra and its Applications}, vol. 435, no. 3, pp. 641-658, 2011.

%
%
\bibitem{tsvdnorm}
Z. Zhang, G. Ely, S. Aeron, N. Hao, and M. Kilmer, Novel methods for multilinear data completion and de-noising based on tensor-SVD. {\it Proceedings of the IEEE Conference on Computer Vision and Pattern Recognition}, 2014.
%
\bibitem{alm}
Z. Lin, M. Chen, Y. Ma. The augmented Lagrange multiplier method for exact recovery of corrupted low-rank matrices. {\it Technical Report UILUENG-09-2215}, UIUC, 2009.


\bibitem{ukbench}
D.~Nister, H.~Stewenius. Scalable recognition with a vocabulary tree. {\it Proceedings of the IEEE Conference on Computer Vision and Pattern Recognition}. 2006.

\bibitem{haming}
H.~Jegou, M.~Douze, and C.~Schmid. Hamming embedding and weak geometric consistency for large scale image search. {\it European conference on computer vision}, 2008.

\bibitem{three}
R. Arandjelovi$\acute{c}$, A. Zisserman. Three things everyone should know to improve object retrieval. {\it Proceedings of the IEEE Conference on Computer Vision and Pattern Recognition}, 2012.

\bibitem{negative}
H. J$\acute{e}$gou, O. Chum. Negative evidences and co-occurences in image retrieval: The benefit of PCA and whitening. {\it European conference on computer vision}, 2012.


\bibitem{convfeature}
A. Babenko, V. Lempitsky. Aggregating local deep features for image retrieval. {\it Proceedings of the IEEE International Conference on Computer Vision }, 2015.


\bibitem{spatial}
W. Zhou, Y. Lu, H. Li, Q. Tian. Spatial coding for large scale partial-duplicate web image search. {\it Proceedings of the ACM international conference on Multimedia}, 2010.

\bibitem{bagofconv}
E. Mohedano, K. Mcguinness K, N. E. O'Connor, A. Salvador, F. Marqu$\acute{e}$s, X. Gir$\acute{o}$-i-Nieto. Bags of Local Convolutional Features for Scalable Instance Search. {\it Proceedings of the ACM  International Conference on Multimedia Retrieval}, 2016.

\bibitem{coindexing}
S. Zhang, M. Yang, X. Wang, Y. Lin, Q. Tian. Semantic-Aware Co-Indexing for Image Retrieval. {\it IEEE Trans. on Pattern Analysis and Machine Intelligence}, vol 37, no. 12, pp. 2573-2587, 2015.

\bibitem{mulindex}
A. Babenko, V. Lempitsky. The Inverted Multi-Index. {\it IEEE Trans. on Pattern Analysis and Machine Intelligence}, vol 37, no. 6, pp. 1247-1260, 2015.

\bibitem{innerproduct}
B. Neyshabur, N. Srebro. On Symmetric and Asymmetric LSHs for Inner Product Search. {\it Proceedings of the international conference on machine learning}, 2015.

\bibitem{LSH}
A. Gionis, P. Indyk, R. Motwani. Similarity search in high dimensions via hashing. {\it Proceedings of the VLDB Conference}, 1999.

\bibitem{mulhash}
Y. Hao, T. Mu, R. Hong, M. Wang, N. An, J. Y. Goulermas. Stochastic Multiview Hashing for Large-Scale Near-Duplicate Video Retrieval. {\it  IEEE Trans. on Multimedia}, vol 19, no. 1, pp. 1-14, 2016.


\bibitem{spectralhash}
Y. Weiss, A. Torralba, R. Fergus. Spectral hashing. {\it Advances in neural information processing systems}, 2009.


\bibitem{hashreview}
J. Wang, W. Liu, S. Kumar, S. F. Chang. Learning to hash for indexing big data¡ªa survey. {\it Proceedings of the IEEE}, 2016.

\bibitem{SSC}
E. Elhamifar, R. Vidal. Sparse subspace clustering: Algorithm, theory, and applications. {\it IEEE Trans. on Pattern Analysis and Machine Intelligence}, vol 35, no. 11, pp. 2765-2781, 2013.

\bibitem{msc}
C. Zhang, H. Fu, S. Liu, G. Liu, X. Cao. Low-rank tensor constrained multiview subspace clustering. {\it Proceedings of the IEEE International Conference on Computer Vision}, 2015.

\bibitem{ts}
O. Semerci, N. Hao, M. E. Kilmer, et al. Tensor-based formulation and nuclear norm regularization for multienergy computed tomography. {\it IEEE Transactions on Image Processing}, vol 24, no.4, pp.1678-1693, 2014.

\bibitem{tproduct}
M. E. Kilmer, K. Braman, N. Hao, R. C. Hoover. Third-order tensors as operators on matrices: A theoretical and computational framework with applications in imaging. {\it SIAM Journal on Matrix Analysis and Applications}, vol 34, no. 1, pp. 148-172, 2013.

\bibitem{mulllindex}
X. Chen, J. Wu, S. Sun, Q. Tian. Multi-Index Fusion via Similarity Matrix Pooling for Image Retrieval. {\it IEEE International Conference on Communications}, 2017.

\bibitem{calembed}
W. Zhou, H. Li, Q. Tian, J. Sun. Collaborative Index Embedding for Image Retrieval. {\it IEEE Trans. on Pattern Analysis and Machine Intelligence}, doi=10.1109/TPAMI.2017.2676779, 2017.


\bibitem{market}
L. Zheng, L. Shen, L. Tian, S. Wang, J. Wang, Q. Tian. Scalable person re-identification: A benchmark. {\it  Proceedings of the IEEE International Conference on Computer Vision}, 2015.
\bibitem{reidreview}
L. Zheng, Y. Yang, A. G. Hauptmann. Person Re-identification: Past, Present and Future. {\it arXiv preprint arXiv:1610.02984}, 2016.


\bibitem{lookba}
I. B. Barbosa, M. Cristani, B. Caputo, A. Rognhaugen, T. Theoharis. Looking beyond appearances: Synthetic training data for deep cnns in re-identification. {\it arXiv preprint arXiv:1701.03153}, 2017.

\bibitem{NSH}
Z. Chen, J. Lu, J. Feng, J. Zhou. Nonlinear Sparse Hashing. {\it IEEE Trans. on Multimedia}, doi=10.1109/TMM.2017.2705918, 2017.

\bibitem{rerank}
Z. Zhong, L. Zheng, D. Cao, S. Li. Re-ranking Person Re-identification with k-reciprocal Encoding. {\it Proceedings of the IEEE Conference on Computer Vision and Pattern Recognition}, 2017.
\bibitem{NULL}
L. Zhang, T. Xiang, S. Gong. Learning a Discriminative Null Space for Person Re-identification. {\it IEEE Conference on Computer Vision and Pattern Recognition}. 2016.
\bibitem{S-LSTM}
R. R. Varior, B. Shuai, J. Lu, D. Xu, G. Wang. A Siamese Long Short-Term Memory Architecture for Human Re-identification. {\it European conference on computer vision}, 2016.
\bibitem{Gate-CNN}
R. R. Varior, B. Shuai, M. Haloi, G. Wang. Gated siamese convolutional neural network architecture for human reidentification. {\it European conference on computer vision}, 2016.
\bibitem{SCSP}
D. Chen, Z. Yuan, B. Chen, and N. Zheng. Similarity learning with spatial constraints for person re-identification. {\it IEEE Conference on Computer Vision and Pattern Recognition}, 2016.
\bibitem{SSDAL}
C. Su, S. Zhang, J. Xing, W. Gao, and Q. Tian. Deep attributes driven multi-camera person re-identification. {\it European conference on computer vision}, 2016.
\bibitem{uniting}
Z. Liu, H. Li, W. Zhou, R. Hong, Q, Tian. Uniting keypoints: Local visual information fusion for large-scale image search. {\it IEEE Trans. on Multimedia}, vol 17, no. 4, pp. 538-548, 2015.
\end{thebibliography}
\end{document}